\documentclass[preprint,12pt]{elsarticle}

\journal{Data \& Knowledge Engineering}

\usepackage{amsmath,amsfonts}
\usepackage{amssymb, amsthm} % amsthm is needed for 

\usepackage{array}

\newcommand{\algCmnd}[1]{\hfill \textcolor{mygray}{{\small /* \texttt{#1} */}}}

\usepackage[font=footnotesize,labelfont=bf,textfont=footnotesize]{subfig}
\usepackage{textcomp}
\usepackage{stfloats}
\usepackage{url}
\usepackage{verbatim}
\usepackage{graphicx}
\newtheorem{example}{Example}
\newtheorem{definition}{Definition}
\newtheorem{theorem}{Theorem}
\usepackage[hidelinks]{hyperref}
\usepackage{xspace}
\usepackage[utf8]{inputenc}
\usepackage{balance}
\usepackage{amsmath}  % Load amsmath first
\usepackage{booktabs}
\usepackage{caption}
\usepackage{colortbl}
\usepackage{xcolor}    % For adding color
\usepackage{graphicx}
\usepackage{tikz}
\usetikzlibrary{patterns}
\usepackage{siunitx}   % For better number formatting
\usepackage{multirow}
\usepackage{footmisc}
\usepackage{listings}
\usepackage{commath}
\usepackage{epstopdf}
\usepackage{url}
\usepackage{float}
\usepackage{mathrsfs}
\usepackage{bbm}
\usepackage{fancyvrb}
\usepackage[linesnumbered,ruled,vlined]{algorithm2e}
\usepackage{algpseudocode}
\usepackage{mathtools}
\usepackage{enumitem}
\usepackage{array}
\usepackage{arydshln}
\usepackage{cleveref}

\usepackage{float}

\usetikzlibrary{shapes, arrows, positioning}

\definecolor{mygreen}{rgb}{0,0.6,0}
\definecolor{myred}{rgb}{0.6,0,0}
\definecolor{mygray}{rgb}{0.5,0.5,0.5}
\definecolor{mymauve}{rgb}{0.58,0,0.82}
\definecolor{myblue}{rgb}{0,0,1}

\lstnewenvironment{VerbatimText}[1][]{
    
    \lstset{fancyvrb=true,basicstyle=\footnotesize,captionpos=b,xleftmargin=2em,#1}
}{}

\usepackage{tikz}
\usetikzlibrary{positioning, arrows.meta}

\definecolor{highlight}{rgb}{0.9, 0.9, 0.5}

\newcommand{\Dom}[1]{\ensuremath{Dom}(#1)}

\DeclareMathAlphabet\mathbfcal{OMS}{cmsy}{b}{n}

\newcommand{\E}{{\tt \mathbb{E}}}
\definecolor{ReviewerOneColor}{RGB}{240, 110, 0}   % Orange
\definecolor{ReviewerTwoColor}{RGB}{0, 139, 109}   % Dark Cyan
\definecolor{ReviewerThreeColor}{RGB}{45, 85, 205}
\definecolor{MetaReviewerColor}{RGB}{142, 69, 133} % Plum

\definecolor{darkred}{rgb}{0.55, 0.0, 0.0}

\newcommand{\nit}[1]{\textit{#1}}

\newcommand{\TPR}{\text{TPR}\xspace}
\newcommand{\FPR}{\text{FPR}\xspace}

\newcommand{\threshold}{\theta}
\newcommand{\PR}{\text{PR}\xspace}

\newcommand{\EOD}{\text{EOD}\xspace}
\newcommand{\EO}{\text{EO}\xspace}
\newcommand{\DP}{\text{DP}\xspace}

\newcommand{\plabel}{\hat{Y}\xspace}
\newcommand{\tlabel}{Y\xspace}

\newcommand{\weightD}{\alpha}
\newcommand{\fBias}{\nit{bias}\xspace}
\newcommand{\fRisk}{\nit{risk}\xspace}
\newcommand{\fMetric}{\Phi\xspace}

\newcommand{\fProblem}{\text{FairScore}\xspace}

\newcommand{\sourceD}{P\xspace}
\newcommand{\targetD}{Q\xspace}
\newcommand{\allCoupling}{\Pi\xspace}
\newcommand{\coupling}{\pi\xspace}

\usepackage{bbm}

\newcommand{\relation}{R}
\newcommand{\allTuples}{\mathcal{R}}
\newcommand{\tuple}[1]{#1}
\newcommand{\rPair}[1]{#1}

\newcommand{\schema}{\ensuremath{\mathcal{S}}\xspace}

\newcommand{\matcher}{f\xspace}

\newcommand{\equivPairs}{E}
\newcommand{\ScoreFunc}{s}
\newcommand{\probD}{P}
\newcommand{\senAtt}{A}
\newcommand{\Quantile}{\mathbb{Q}}
\newcommand{\CDF}{\mathbb{F}}
\newcommand{\Dmap}{T}

\newcommand{\minor}{a\xspace}
\newcommand{\major}{b\xspace}

\definecolor{Teal}{HTML}{1B9E77}    % Green-ish alternative
\definecolor{SkyBlue}{HTML}{56B4E9} % Blue alternative
\definecolor{Vermillion}{HTML}{D55E00} % Red alternative

\newcommand{\ditto}{\texttt{DITTO}\xspace}
\newcommand{\hiergat}{\texttt{HierGAT}\xspace}
\newcommand{\dmatcher}{\texttt{DeepMatch}\xspace}
\newcommand{\hmatcher}{\texttt{HierMatch}\xspace}
\newcommand{\EMT}{\texttt{EMTransform}\xspace}

\newcommand{\walamz}{\texttt{WAL-AMZ}\xspace} % walmart-amazon
\newcommand{\amzgog}{\texttt{AMZ-GOO}\xspace} % amazon- google
\newcommand{\DBLPGogS}{\texttt{DBLP-GOO}\xspace} % dblp - google scholar
\newcommand{\DBLPACM}{\texttt{DBLP-ACM}\xspace} % dblp- acm

\newcommand{\itunamz}{\texttt{ITU-AMZ}\xspace} % itunes-amazon
\newcommand{\FodorZag}{\texttt{FOD-ZAG}\xspace} % fodors - zagat
\newcommand{\Beer}{\texttt{BEER}\xspace} % Beer
\newcommand{\boxtheorem}{\hfill $\blacksquare$\vspace{2mm}}

\newcommand{\ignore}[1]{}

\definecolor{black}{rgb}{0,0,0}
\definecolor{grey}{rgb}{0.8,0.8,0.8}
\definecolor{red}{rgb}{1,0,0}
\definecolor{green}{rgb}{0,1,0}
\definecolor{darkgreen}{rgb}{0,0.5,0}
\definecolor{darkpurple}{rgb}{0.5,0,0.5}
\definecolor{darkdarkpurple}{rgb}{0.3,0,0.3}
\definecolor{blue}{rgb}{0,0,1}
\definecolor{shadegreen}{rgb}{0.95,1,0.95}
\definecolor{shadeblue}{rgb}{0.95,0.95,1}
\definecolor{shadered}{rgb}{1,0.85,0.85}
\definecolor{shadegrey}{rgb}{0.85,0.85,0.85}
\definecolor{oddRowGrey}{rgb}{0.80,0.80,0.80}
\definecolor{evenRowGrey}{rgb}{0.85,0.85,0.85}
\definecolor{lightpurple}{rgb}{0.88,1.0,1.0}

\newcommand{\indep}{\mbox{$\perp\!\!\!\perp$}}

\newcommand{\att}{A}

\newcommand{\RNum}[1]{\uppercase\expandafter{\romannumeral #1\relax}}
\newtheorem{defn}{Definition}[section]

\newtheorem{col}[defn]{Colollary}

\newcommand{\proj}[1]{{\Pi}}
\newcommand{\sel}[1]{{\sigma}}

\newcommand{\cut}[1]{}
\newcommand{\eat}[1]{}

\SetKwInput{KwInput}{Input}
\SetKwInput{KwOutput}{Output}

\ExplSyntaxOn
  \cs_new_eq:NN \calc \fp_eval:n
\ExplSyntaxOff

\definecolor{cbgreen}{RGB}{0,158,115} % Bluish-green
\definecolor{cbred}{RGB}{213,94,0} % Vermillion, a reddish-orange that is often recommended for being colorblind-friendly

\definecolor{DodgerBlue}{RGB}{30, 144, 255}

\usepackage[normalem]{ulem} % This package provides the \sout{} command

\newcommand{\ccalib}{\texttt{C-Calib}\xspace}
\newcommand{\calib}{\texttt{Calib}\xspace}
\newcommand{\cthreshold}{\gamma}

\begin{document}

\begin{frontmatter}

\title{Reducing Biases in Record Matching Through Scores Calibration}

\author[1]{Mohammad Hossein Moslemi\corref{cor1}}
\ead{mohammad.moslemi@uwo.ca}

\author[1]{Mostafa Milani}
\ead{mostafa.milani@uwo.ca}

\cortext[cor1]{Corresponding author.}

\address[1]{Department of Computer Science, Western University, London, Ontario, Canada}

\begin{abstract}
Record matching models typically output a real-valued matching score that is later consumed through thresholding, ranking, or human review. While fairness in record matching has mostly been assessed using binary decisions at a fixed threshold, such evaluations can miss systematic disparities in the \emph{entire} score distribution and can yield conclusions that change with the chosen threshold.

We introduce a threshold-independent notion of \emph{score bias} that extends standard group-fairness criteria---demographic parity (\DP), equal opportunity (\EO), and equalized odds (\EOD)---from binary outputs to score functions by integrating group-wise metric gaps over all thresholds. Using this metric, we empirically show that several state-of-the-art deep matchers can exhibit substantial score bias even when appearing fair at commonly used thresholds.

To mitigate these disparities without retraining the underlying matcher, we propose two model-agnostic post-processing methods that only require score evaluations on an (unlabeled) calibration set. \calib targets \DP by aligning minority/majority score distributions to a common Wasserstein barycenter via a quantile-based optimal-transport map, with finite-sample guarantees on both residual \DP bias and score distortion. \ccalib extends this idea to label-dependent notions (\EO/\EOD) by performing barycenter alignment conditionally on an estimated label, and we characterize how its guarantees depend on both sample size and label-estimation error.

Experiments on standard record-matching benchmarks and multiple neural matchers confirm that \calib and \ccalib substantially reduce score bias with minimal loss in accuracy.
\end{abstract}


\end{frontmatter}

\section{Introduction}\label{sec:intro}

Record matching is the task of identifying records from one or more datasets that refer to the same real-world entity. It is important both as a standalone task—such as matching social media profiles across different platforms or identifying duplicate patient health records in healthcare systems—and as a component of broader data processes like data integration and data cleaning.

Record matching is often formulated as a binary classification problem, where pairs of records are labeled as either ``match'' or ``non-match.'' However, in practice, most record matching methods output a {\em matching score} rather than a direct binary label. This score indicates the likelihood of a match: higher values suggest stronger evidence, while lower values imply that the records likely refer to different entities. A binary decision can be derived by applying a threshold to the score.

Having a matching score, rather than only a binary outcome, is important from both application and technical perspectives. 

From an application perspective, scores allow adjusting the threshold to balance precision and recall. Lower thresholds increase recall by identifying more matches, but may also increase false positives (FP), leading to incorrect record merges and potential data loss. Higher thresholds reduce FPs but risk missing true matches, which can be critical in domains where missing links between records may result in costly consequences. Matching scores also reflect confidence levels and can be used directly in applications. For example, instead of converting scores to binary decisions, one can use them to rank candidate records and return top matches for a given input. This is useful when showing the most likely matches is more informative than a simple yes-or-no answer.

From a technical perspective, matching scores support the development of more flexible and advanced methods. For example, active learning approaches can use scores to identify uncertain cases—those with values near the decision threshold—and prioritize them for manual labeling to improve the model~\cite{kasai2019low,meduri2020comprehensive}. This idea also extends to human-in-the-loop systems, where scores help determine which record pairs should be reviewed by a human. Instead of treating all pairs equally, the system can focus on those that are harder to classify automatically~\cite{sarawagi2002interactive,elmagarmid2014nadeef}. Another technical reason for using scores instead of a binary outcome is that many state-of-the-art matching techniques, especially deep learning models, naturally produce scores through their architecture—for example, by applying a sigmoid activation to the output of a final linear layer, resulting in a value between 0 and 1 that reflects the predicted match likelihood. Because of these advantages, matching scores are more widely used than hard binary decisions in both research and practical systems.

In this paper, we study fairness in record matching by focusing on biases in matching scores and propose methods to reduce these biases. Fairness in record matching has received growing attention in recent years, as studies have shown that biases in matching can disproportionately harm minority groups and lead to unfair outcomes~\cite{shahbazi2024fairness, efthymiou2021fairer, nilforoushan2022entity, shahbazi2023through}. However, prior work has mainly treated record matching as a binary classification task. In contrast, biases in matching scores are more complex and harder to detect and reduce. A score may appear fair under one threshold but produce biased outcomes under another. Moreover, in many applications where scores are used directly—rather than converted to binary decisions—differences in score quality across groups can lead to lower accuracy for minorities and unfair treatment. The following example illustrates how biases in matching scores can be more subtle and problematic than in binary outcomes.

\begin{figure}[H]
    \centering
    \subfloat[TPR vs Matching Threshold]{\label{fig:tpr}
        \includegraphics[width=0.45\textwidth]{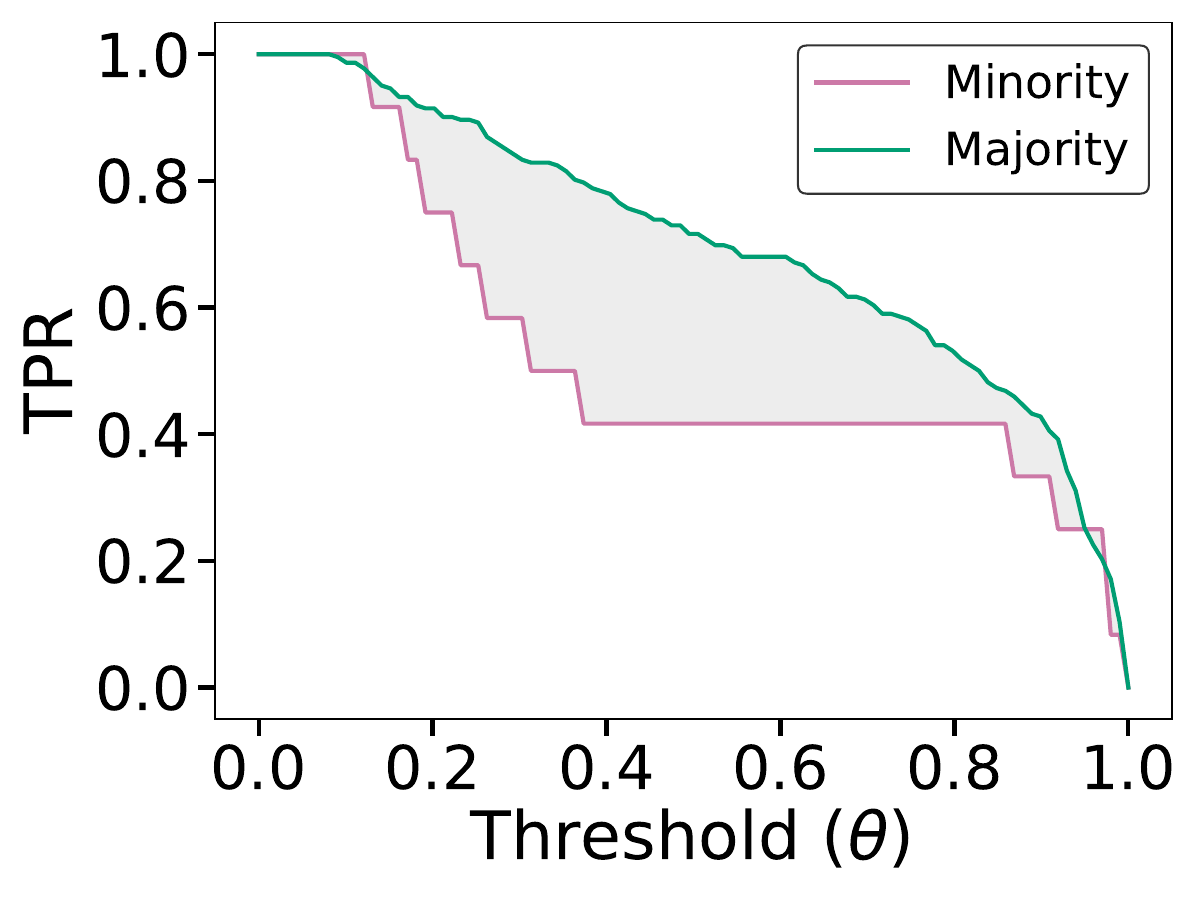}}
    \hfill
    \subfloat[{AUC}]{\label{fig:auc-two}
        \includegraphics[width=0.45\textwidth]{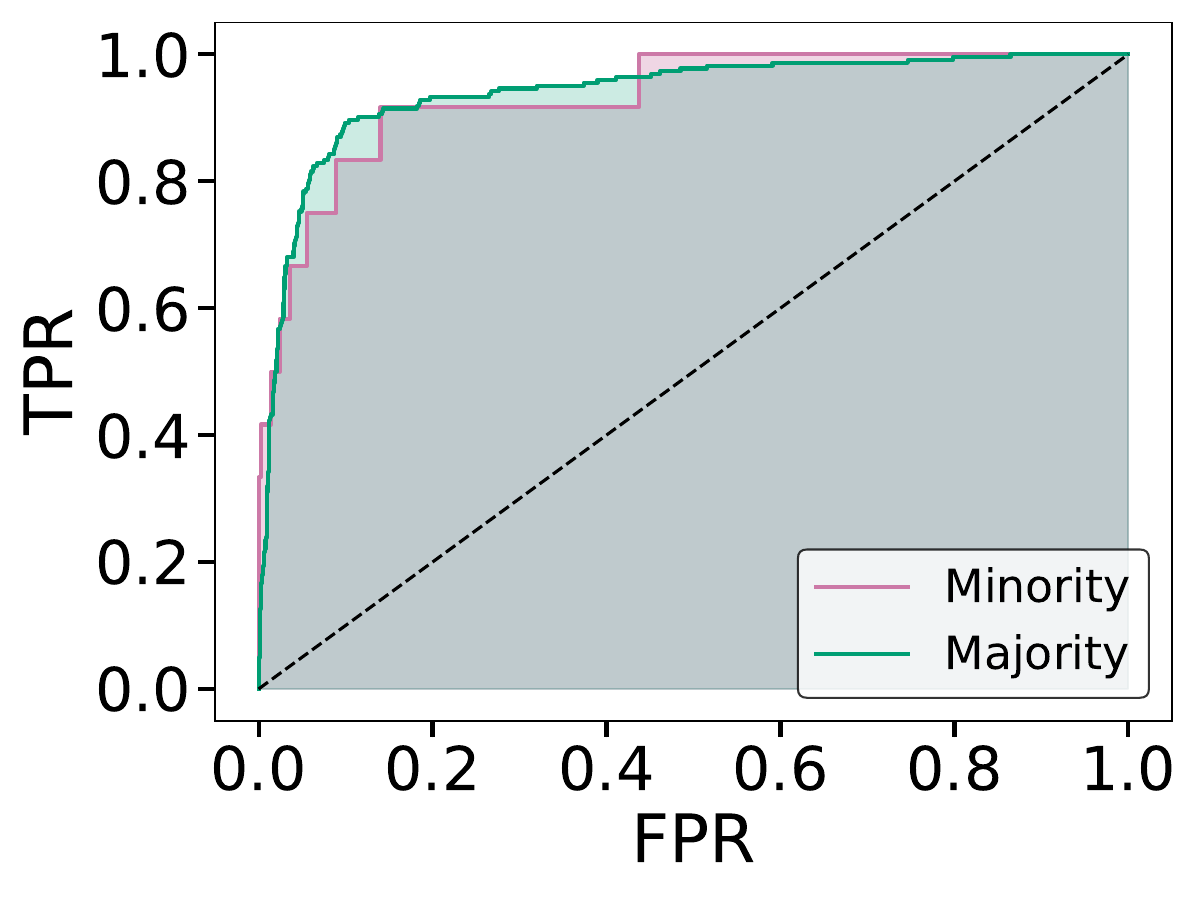}}
    \caption{\normalsize Figure~\ref{fig:tpr} shows variation in \TPR across thresholds. The ROC of \hmatcher~\cite{fu2021hierarchical} (Figure~\ref{fig:auc-two}) shows that the AUC is nearly the same for both groups, with 93.33\% for the minority group and 93.94\% for the majority. However, there is a noticeable difference in performance at specific thresholds.
}
    \label{fig:intro}
\end{figure}

\begin{example}\em \label{ex:intro}
Figure~\ref{fig:intro} shows the performance of \hmatcher~\cite{fu2021hierarchical}, a state-of-the-art record matching method, on \DBLPACM---a benchmark of publication records with potential duplicates referring to the same paper.
We compare the method’s performance on records authored by females as the minority group, versus all other records as the majority group. Figure~\ref{fig:tpr} reports the true positive rate (\TPR) for each group and for varying matching thresholds. The gap in \TPR\ near mid-range thresholds reflects bias in binary outcomes. At thresholds near 0 or 1, these gaps disappear, suggesting no bias at those extremes. This demonstrates that score biases depend on the threshold and must be evaluated across the full score range by comparing score distributions across groups.

Figure~\ref{fig:auc-two} compares the ROC curves of the matcher for the minority and majority groups. Each ROC curve summarizes performance across all thresholds by plotting \TPR\ versus \FPR, and the area under the curve (AUC) is a common threshold-independent measure of score quality. A standard fairness check is to compare AUC across groups.

In this example, however, the group AUC values are almost identical even though the ROC curves differ: one group has higher \TPR\ in some regions of \FPR\ and lower \TPR\ in others. These differences can matter at the thresholds used in practice, but they can cancel out when aggregated into a single AUC number. This illustrates why AUC-based, threshold-independent comparisons can still miss meaningful score bias.
\boxtheorem
\end{example}

The most closely related work to fairness in matching scores is the literature on fairness in classification scores for binary classification tasks (e.g.,~\cite{kallus2019fairness, yang2023minimax, vogel2021learning}). These approaches primarily rely on AUC-based measures, defining bias as the gap in AUC between majority and minority groups. However, such metrics can be misleading~\cite{kwegyir2023misuse}, as shown in Example~\ref{ex:intro}. Moreover, our setting differs in that matching scores are often skewed and context-specific, unlike the general case of binary classification. This distinction motivates the need for dedicated techniques to assess and mitigate bias in matching scores.

Fairness in matching scores is closely related to fairness in both \emph{regression} and \emph{ranking}, since all three settings use a real-valued score that may exhibit systematic differences across protected groups.

In fair regression, the output is a continuous prediction, and many approaches aim to reduce group disparities by constraining or post-processing these numerical outputs~\cite{komiyama2018general,agarwal2019fair}. A particularly relevant line of work treats predictions as samples from group-wise distributions and then aligns these distributions via post-processing (e.g., using optimal transport and Wasserstein barycenters)~\cite{chzhen2020fair}. This is closely aligned with our approach: we view matching scores for the minority and majority as two empirical distributions and post-process them to a common target distribution (a barycenter), thereby reducing threshold-independent score disparities while keeping score distortion small.

Fair ranking is also relevant because ranking pipelines rely on scores that are then consumed through ordering (e.g., top-$k$ selection). Many fair-ranking methods seek to enforce group-level representation or exposure constraints in the top-ranked results (e.g.,~\cite{yang2019balanced, celis2018ranking, celis2020interventions}). These methods highlight how biases in scores can translate into biased downstream decisions, and they provide a complementary perspective to our focus on fairness of the scores themselves.

More broadly, there are many other techniques in both fair regression and fair ranking, including in-processing constraints and alternative post-processing strategies~\cite{komiyama2018general,agarwal2019fair,yang2019balanced,celis2018ranking,celis2020interventions}. In this work, we primarily build on the distributional-alignment perspective from fair regression, since it naturally matches our goal of reducing disparities in the \emph{entire score distribution}.

At the same time, record matching introduces important differences with fair regression and fair ranking. Matching scores are typically bounded in $[0,1]$, often concentrated near 0 and 1, and are used through thresholding, ranking, and human-in-the-loop review.
More importantly, record matching is a setting where \emph{both} the score and the induced binary decision are meaningful outcomes, so fairness notions naturally include label-dependent criteria such as equal opportunity (\EO) and equalized odds (\EOD). These criteria require aligning error rates (e.g., \TPR/\FPR) across groups and therefore depend on the (unknown) true match/non-match label.
This label dependence is a main technical difference from most ranking and regression settings (and from many of their fairness objectives), where the fairness constraint is typically defined without conditioning on ground-truth labels.
Thus, while distributional alignment ideas can directly target threshold-independent score disparity, extending them to \EO/\EOD is nontrivial when labels are unavailable at calibration time. This motivates our \emph{conditional calibration} approach, which conditions distributional alignment on an estimated label so it can target \EO/\EOD in settings where true labels are unavailable.

Our contributions are as follows.

\noindent\textbf{(1) Score-bias formulation for record matching.} We introduce a threshold-independent bias measure that extends standard notions such as demographic parity (\DP) and equalized odds (\EOD) from binary outcomes to score functions. It captures cumulative performance differences between groups over the full threshold range. For example, in Figure~\ref{fig:tpr}, the \DP score bias corresponds to the area between group-wise \TPR curves across thresholds.

\noindent\textbf{(2) Barycenter-based score calibration for \DP.} Building on distributional calibration ideas from the fair regression literature, we propose a model-agnostic post-processing algorithm that maps group-wise score distributions to a common Wasserstein barycenter~\cite{chzhen2020fair, miroshnikov2022wasserstein}. This reduces \DP score bias while keeping score distortion small; moreover, our theoretical guarantees for the \DP case build directly on theory developed in the fair regression literature.

\noindent\textbf{(3) Conditional calibration for label-dependent notions (\EO/\EOD).} We introduce \emph{conditional calibration}, which applies barycenter-based alignment \emph{within strata defined by an estimated label} (since ground-truth labels may be unavailable at calibration time). This extends score calibration beyond \DP to address label-dependent disparities such as \EO and \EOD, and we extend the theoretical analysis to this setting by characterizing how \EO/\EOD fairness depends on both sample size and label-estimation error.

\noindent\textbf{(4) Empirical evaluation.} Experiments across standard record-matching benchmarks and state-of-the-art matchers show that the proposed methods substantially reduce score bias with minimal impact on accuracy.

This paper is organized as follows. Section~\ref{sec:background} presents background on record matching, notation, and existing fairness metrics. Section~\ref{sec:Problem} introduces our proposed fairness metric and formalizes the problem. Sections~\ref{sec:solv} and~\ref{sec:partial} describe two bias reduction algorithms. Section~\ref{sec:experiments} reports experimental results. Section~\ref{sec:rw} reviews related work and Section~\ref{sec:conclusionfinal} concludes with a summary and future directions.

\section{Preliminaries} \label{sec:background}

A relational schema $\schema$ consists of attributes $\att_1, \dots, \att_m$, where each attribute $\att_i$ has a domain $\Dom{\att_i}$ for $i \in [1, m]$. A record (or tuple) $\tuple{r}$ over $\schema$ is an element of the Cartesian product $\Dom{\att_1} \times \dots \times \Dom{\att_m}$, representing the space of all possible records, denoted by $\allTuples$. The value of attribute $\att_i$ in record $\tuple{r}$ is written as $\tuple{r}[\att_i]$. A relation (or instance) $\relation$ over $\schema$ is a subset of $\allTuples$ containing a set of records.

\subsection{Record Matching} \label{sec:record-matching}

Given two relations, $\relation_1$ and $\relation_2$ with schemas $\schema_1$ and $\schema_2$, \emph{record matching} aims to identify a subset $\equivPairs$ of record pairs in $\relation_1 \times \relation_2$ that refer to the same real-world entities. These are called {\em equivalent pairs}, while all others are {\em non-equivalent}. Without loss of generality and for simplicity, we assume $\relation_1 = \relation_2$. A record matcher, or simply a matcher, is a binary classifier $\matcher: \allTuples^2 \rightarrow \{0,1\}$ that assigns a label of $1$ (match) or $0$ (non-match) to pairs of records from a relation $\relation$ with schema $\schema$. The goal is to correctly label equivalent pairs as matches and non-equivalent pairs as non-matches. A matcher $f$ is often defined using a matching score function $\ScoreFunc: \allTuples^2 \to [0,1]$ and a threshold $\threshold \in [0,1]$. It labels a pair $\rPair{p}$ as a match if $\ScoreFunc(\rPair{p}) \geq \threshold$, i.e., $f(\rPair{p}) = \mathbbm{1}_{s(\rPair{p}) \geq \threshold}$. Performance metrics for $\matcher$ are formally defined with respect to a joint probability distribution $\probD$ over a random variable $X$ (representing pairs of records from $\schema$) and a binary variable $Y$ indicating the true label. The positive rate (\PR), true positive rate (\TPR), and false positive rate (\FPR) for $\ScoreFunc$ and $\threshold$ are defined as $\PR(\ScoreFunc, \threshold) = \probD(\ScoreFunc(X) \ge \threshold)$, $\TPR(\ScoreFunc, \threshold) = \probD(\ScoreFunc(X) \ge \threshold \mid \tlabel = 1)$, and $\FPR(\ScoreFunc, \threshold) = \probD(\ScoreFunc(X) \ge \threshold \mid \tlabel = 0)$. We use $\plabel$ to refer to $\matcher(X)$ and, when clear from context, write $\att_i$ instead of $A$ and $X[\att_i]$ for attribute values. Other metrics such as precision, recall, and F1 score are defined similarly. The AUC for $\ScoreFunc$ can be written as $\int_0^1 \TPR(\ScoreFunc, \FPR^{-1}(\ScoreFunc, u))\, du$.

\begin{table}
  \centering
  \setlength\tabcolsep{2 pt}
 \begin{tabular}{l l}
  \toprule
    Symbol              & Description \\
    \midrule
    $\relation,\schema$   & Relation and schema\\
    $r,p$   & Record and (record) pair\\
    $A_1,A_2,...$   & Attributes\\
    $A$ & Binary sensitive (protected) attribute\\
    $\allTuples$   & All possible records\\ 
    $\Dom{A_i}$ & Domain of an attribute $A_i$\\ 
    $a,b \in  \Dom{A}$  & Minority and majority \\      
    $P(X)$ & Probability distributions with random variable $X$\\
    $\fMetric$ & Performance metric\\
    $\tilde{P}$ & Wasserstein barycenter\\
    $s,\theta$ & Matching score function and matching threshold\\
    \bottomrule
    \end{tabular}
  \caption{Summary of notation and symbols.}
  \label{tab:symbols}
\end{table}

\subsection{Fairness Measure in Record Matching}\label{sec:bias-matching}

To study fairness in record matching, we assume records contain a sensitive attribute $\senAtt \in \schema$, such as gender or ethnicity, which defines majority and minority groups. For simplicity, we consider $\senAtt$ to be binary, with $\Dom{\senAtt} = \{\minor, \major\}$, where records with value $\minor$ belong to the minority group (e.g., female) and those with $\major$ to the majority group (e.g., male). A record pair $\rPair{p} = (\tuple{r}_1, \tuple{r}_2)$ is considered a minority pair, denoted by $\rPair{p}[\senAtt] = \minor$, if either record belongs to the minority group, i.e., if $\tuple{r}_1[A] = \minor$ or $\tuple{r}_2[A] = \minor$; otherwise, $\tuple{r}_1[A] = \major$ and $\tuple{r}_2[A] = \major$ and it is a majority pair, which we denote by $\rPair{p}[\senAtt] = \major$. This definition ensures that any pair involving a minority record is treated as a minority pair, reflecting that mismatches may disproportionately affect minority records.

\subsubsection{Fairness in Binary Matching} \label{sec:fairness-binary-match}

Fairness in record matching, treated as a binary classification task, has been extensively studied in~\cite{shahbazi2023through}. Several fairness definitions are applied to assess bias, following principles similar to those used in general classification tasks. These definitions examine the relationship between the sensitive attribute $\rPair{p}[\senAtt]$ and the matcher’s output $f(\rPair{p})$ for any record pair $\rPair{p}$. When a matcher is defined by a score function $s$ and a threshold $\theta$, these fairness metrics are threshold-dependent; we make this explicit using a subscript $\theta$ (e.g., $\DP_\theta$).

Each definition compares a {\em performance metric} $\fMetric$, such as \TPR, \FPR, or \PR, across sensitive groups, denoted $\fMetric_g$ for $g \in \{\minor, \major\}$. \emph{Demographic parity}, or \emph{statistical parity}, requires the positive rate ($\PR$) to be independent of the sensitive attribute, i.e., $\plabel \indep \senAtt$. Formally, $\PR_a(s,\theta) = \PR_b(s,\theta)$, where $\PR_g(s,\theta) = \probD(\plabel = 1 \mid \senAtt = g)$ and $\plabel = \mathbbm{1}_{s(X)\ge \theta}$. The corresponding demographic parity difference is $\DP_\theta = |\PR_a(s,\theta) - \PR_b(s,\theta)|$.

In contrast, definitions such as \emph{equalized odds} and \emph{equal opportunity} condition on equivalence to evaluate fairness, requiring $\plabel \indep \senAtt \mid \tlabel$. \emph{Equal opportunity (EO)} requires $\TPR_a(s,\theta) = \TPR_b(s,\theta)$, where $\TPR_g(s,\theta) = \probD(\plabel = 1 \mid \senAtt = g, \tlabel = 1)$. The corresponding difference is $\EO_\theta = |\TPR_a(s,\theta) - \TPR_b(s,\theta)|$.

\emph{Equalized odds (EOD)} extends this further by requiring both $\TPR$ and $\FPR$ to be equal across groups. A matcher satisfies \EOD at threshold $\theta$ if $\TPR_a(s,\theta) = \TPR_b(s,\theta)$ and $\FPR_a(s,\theta) = \FPR_b(s,\theta)$, where $\FPR_g(s,\theta) = \probD(\plabel = 1 \mid \senAtt = g, \tlabel = 0)$. We quantify this via $\EOD_\theta = |\TPR_a(s,\theta) - \TPR_b(s,\theta)| + |\FPR_a(s,\theta) - \FPR_b(s,\theta)|$.

\subsection{Wasserstein Distance and Barycenter}

The {\em Wasserstein-$\rho$ distance} ($W_\rho$) quantifies the distance between two probability measures as the minimum cost of transporting mass from one distribution to the other. For univariate distributions $\sourceD$ and $\targetD$ over $\mathbb{R}$, it is defined as:
\begin{align} \label{fr:Wp}
W_\rho(\sourceD, \targetD) = \left( \inf_{\coupling \in \allCoupling(\sourceD, \targetD)} \int_{\mathbb{R} \times \mathbb{R}} d(x, y)^\rho \, d\coupling(x, y) \right)^{1/\rho}
\end{align}
Here, $\allCoupling(\sourceD, \targetD)$ denotes the set of all {\em couplings} of $\sourceD$ and $\targetD$, where each coupling $\coupling$ is a joint distribution over $\mathbb{R}^2$ with marginals $\sourceD$ and $\targetD$. A coupling specifies how to assign mass from $\sourceD$ to $\targetD$ while preserving their distributions. Note that the coupling in Eq.~\ref{fr:Wp} is not unique; different couplings can yield different transport costs. The function $d(x, y)$ represents the cost of transporting mass from $x$ to $y$. For the commonly used $W_2^2$, this cost is $d(x, y) = |x - y|^2$~\cite{villani2009optimal}.

This formulation follows the \emph{Kantorovich approach} to optimal transport, which allows mass at a source point to be spread probabilistically across multiple target points. This flexibility ensures the existence of an optimal coupling that minimizes total transport cost. In contrast, the \emph{Monge formulation} seeks a deterministic mapping $\Dmap$ that transports each source point to a single target point. This is expressed as the \emph{pushforward} $\Dmap_\# \sourceD = \targetD$, meaning that applying $\Dmap$ to $\sourceD$ yields $\targetD$. While more intuitive, the Monge approach can be too restrictive or infeasible when the source and target distributions differ significantly in structure.

The {\em barycenter} of a set of distributions is a representative distribution that balances their characteristics. Depending on the metric used, different types of barycenters can be defined. The {\em Wasserstein barycenter} minimizes the average Wasserstein distance to a given set of distributions, yielding a central distribution that optimally reflects transport-based similarity. Given $k$ distributions $\sourceD_1, \sourceD_2, ..., \sourceD_k$ with weights $\weightD_1, \weightD_2, ..., \weightD_k$, the Wasserstein barycenter $\tilde{\sourceD}$ is defined as~\cite{agueh2011barycenters, cuturi2014fast}: $\tilde{\sourceD} = \arg \min_{\sourceD} \sum_{i=1}^k \weightD_i W_\rho^\rho(\sourceD, \sourceD_i)$. For the special case of two distributions $\sourceD_1$ and $\sourceD_2$ using the Wasserstein-2 distance, this reduces to:
\begin{align} \label{fr:WassBary2D}
\tilde{\sourceD} = \arg \min_{\sourceD} \left( \weightD W_2^2(\sourceD_1, \sourceD) + (1 - \weightD) W_2^2(\sourceD_2, \sourceD) \right)
\end{align}

\section{Problem Definition} \label{sec:Problem}

As discussed in Section~\ref{sec:bias-matching}, existing fairness definitions are either threshold-specific or, in the case of AUC-based, threshold-independent but potentially misleading. To address these limitations, we introduce a new fairness metric for matching scores, extending binary classification fairness measures (see Section~\ref{sec:fairness-binary-match}) to evaluate biases in a score function $s$ across all thresholds.

\begin{definition}[Fair matching score] \label{df:fairscore} \em
The bias of a matching score function $\ScoreFunc$ with respect to a performance metric $\fMetric$ is defined as
\begin{align}
\fBias(\ScoreFunc,\fMetric)=\int_0^1 |\fMetric_\major(\ScoreFunc,\threshold) - \fMetric_\minor(\ScoreFunc,\threshold)|\;d\threshold \label{eq:fair}
\end{align}
We say $\ScoreFunc$ is fair if $\fBias(\ScoreFunc,\fMetric) = 0$.
\end{definition}

Definition~\ref{df:fairscore} generalizes fairness notions such as \DP, \EO, and \EOD to score functions. For example, setting $\fMetric = \PR$ recovers \DP for $\ScoreFunc$, referred to as {\em score \DP}. Similarly, $\fMetric = \TPR$ corresponds to score \EO, and using both $\fMetric = \TPR$ and $\fMetric = \FPR$ yields \EOD. For any unfair score function, the value of $\fBias(\ScoreFunc,\fMetric)$ quantifies the score bias. For example, $\fBias(\ScoreFunc,\PR)$ measures the \DP score bias, $\fBias(\ScoreFunc,\TPR)$ captures the \EO score bias, and $\fBias(\ScoreFunc,\TPR) + \fBias(\ScoreFunc,\FPR)$ represents the \EOD score bias. When clear from context, we omit the term ``score'' and simply refer to these as \DP, \EO, or \EOD biases; we reserve subscripts (e.g., $\DP_\theta$, $\EO_\theta$, $\EOD_\theta$) for threshold-specific binary fairness metrics.

Prior work~\cite{shahbazi2023through,shahbazi2024fairness} has shown that many existing record matching methods produce biased matching scores. In Section~\ref{sec:exp-bias}, we demonstrate through experiments on standard benchmarks that several state-of-the-art techniques exhibit substantial bias. To address this, we define the problem of generating fair matching scores and propose a post-processing solution to adjust scores from existing methods.

\begin{definition}[\fProblem] \label{df:fair-problem} \em 
Given a matching score function $\ScoreFunc$ and a performance metric $\fMetric$, the \fProblem problem seeks an optimal fair score function $\ScoreFunc^*$:
\begin{align} \label{fr:optimalScoreF}
\ScoreFunc^* = \arg\min_{\text{fair}\; s'} \fRisk(\ScoreFunc',\ScoreFunc)
\end{align}
\noindent where the risk function is defined as
\begin{align} \label{eq:risk}
\fRisk(\ScoreFunc^*, \ScoreFunc) = \mathbb{E}\big[|\ScoreFunc^*(X) - \ScoreFunc(X)|\big]
\end{align}
and $X$ is a random variable over record pairs.
\end{definition}

In other words, \fProblem aims to find a fair score function that remains close to the original one, minimizing the difference in scores while improving fairness with respect to the chosen performance metric $\fMetric$.

\section{Score Calibration Using Barycenter} \label{sec:solv}

This section introduces our first bias reduction algorithm, \calib (Algorithm~\ref{alg:calibrate}), for the \fProblem problem with $\fMetric = \PR$, where the goal is to remove \DP bias. Given a score function $s$, \calib calibrates it into a new function $\hat{s}$ that is independent of group membership. It does so by computing the Wasserstein barycenter of the minority ($s_\minor$) and majority ($s_\major$) score distributions, balancing both groups while minimizing the risk of $\hat{s}$ (Eq.~\ref{eq:risk}).

We first explain the intuition behind the algorithm and then describe the algorithm step by step with an example. We conclude with a discussion of its theoretical properties.

\subsection{\calib Algorithm} \label{sec:calib}

The \calib algorithm calibrates scores by estimating the Wasserstein barycenter of the score distributions for each group.
It computes the barycenter using \emph{quantile-based barycenter computation} method, a standard approach in optimal transport theory~\cite{peyre2019computational} for univariate distributions. In one dimension, the Wasserstein distance becomes a linear transport problem, making this method a natural choice. This method requires absolutely continuous distributions with shared support--conditions typically met by deep learning models that output real-valued scores (e.g., \([0,1]\)). Alternative methods such as Sinkhorn~\cite{knight2008sinkhorn} are less effective for univariate cases and are primarily used for multivariate distributions and therefore we did not consider them in our work.

Now, consider the univariate distributions of scores for each group, denoted by $\probD_\minor$ and $\probD_\major$. 
The algorithm computes the barycenter distribution $\tilde{\probD}$ of $\probD_\minor$ and $\probD_\major$ using the quantile-based method. 
This approach leverages the \emph{cumulative distribution function (CDF)}, denoted by $\CDF$, and its inverse, the \emph{quantile function} ($\Quantile$), of each distribution. Specifically, the quantile function for group $g\in\{\minor,\major\}$ is defined as $\Quantile_g(p) = \CDF_g^{-1}(p)$, mapping each probability level $p \in [0, 1]$ to its corresponding score in distribution $\probD_g$.

The quantile-based method calculates the barycenter quantile function, $\tilde{\Quantile}(p)$, as a weighted average of the quantile functions of the original distributions:
\[
\tilde{\Quantile}(p) = \alpha \Quantile_\minor(p) + (1 - \alpha) \Quantile_\major(p),
\]
where $\alpha = \probD(\senAtt = \minor)$ and $1 - \alpha = \probD(\senAtt = \major)$. These weights represent the relative contributions of each group distribution to the barycenter. The resulting barycenter distribution $\tilde{\probD}$ integrates characteristics from both original distributions. Finally, the algorithm returns the calibrated score $\hat{s}(\rPair{p})$ derived from the barycenter distribution. Algorithm~\ref{alg:calibrate} summarizes this process.

\begin{algorithm}
\caption{$\calib(D,s,\rPair{p})$}\label{alg:calibrate}
\KwInput{A set of record pairs $D = \{\rPair{p}_1,...,\rPair{p}_{|D|}\}$; a matching score function $s$, a query pair $\rPair{p}$}
\KwOutput{Calibrated score for $\rPair{p}$}

    $\text{scores}_\major \leftarrow []$; $\text{scores}_\minor \leftarrow []$; \label{ln:init-scores} \algCmnd{Initialize group score lists}\\
    \ForEach{$\rPair{p}_i \in D$}{\label{ln:for-each}
    \Indp $g \leftarrow \rPair{p}_i[\senAtt]$;\algCmnd{Identify pair group}\label{ln:get-group}\\
    $\text{append}(\text{scores}_g, s(\rPair{p}_i) + \mathcal{N}(0, \sigma^2))$; \label{ln:append-score}\\
    \Indm }
    $\text{sort}(\text{scores}_\minor)$; 
    $\text{sort}(\text{scores}_\major)$; \label{ln:sort-scores}\\
    $g \leftarrow \rPair{p}[\senAtt]$; \hfill \algCmnd{Identify query pair group}\label{ln:get-query-group}\\
    $\text{pos}_g \leftarrow 0$; \hfill \algCmnd{Initialize insertion point}\label{ln:init-pos}\\
    \While{$\text{pos}_g < |\text{scores}_g|$ \textbf{\em and} $\text{scores}_g[\text{pos}_g] > s(\rPair{p})$}{\label{ln:while-loop}
    \Indp $\text{pos}_g \leftarrow \text{pos}_g + 1$; \label{ln:increment-pos}\\
    \Indm }
    $\alpha \leftarrow \frac{|\text{scores}_\minor|}{|D|}$; $q\leftarrow \frac{\text{pos}_{g}}{|\text{scores}_g|}$;\hspace{-1mm}\algCmnd{Compute quantile}\label{ln:compute-alpha1000}\\
    \eIf{$g=\minor$}{
        $\text{pos}_{\minor}\leftarrow \text{pos}_{g}$;\\
        $\text{pos}_{\major} \leftarrow \left\lceil q\times |\text{scores}_{\major}| \right\rceil$;\algCmnd{Cross-group map}\label{ln:compute-alpha}\\
    }{
        $\text{pos}_{\major}\leftarrow \text{pos}_{g}$;\\
        $\text{pos}_{\minor} \leftarrow \left\lceil q\times |\text{scores}_{\minor}| \right\rceil$;\algCmnd{Cross-group map}\label{ln:compute-alpha}\\
    }
    \Return{$\alpha \times \text{scores}_\minor[\text{pos}_\minor] + (1-\alpha) \times \text{scores}_\major[\text{pos}_\major]$; \label{ln:return-calibrated}}
\end{algorithm}

Before detailing each step in Algorithm~\ref{alg:calibrate}, we clarify a few key points. First, in practice, the distributions \(\probD_\minor\) and \(\probD_\major\) are unknown, which prevents direct use of the quantile method. Instead, the algorithm uses score functions \(s_\minor\) and \(s_\major\), generating i.i.d. samples of scores from randomly selected pairs. To improve robustness, Gaussian noise (commonly known as {jitter}) is added to these samples (Line~\ref{ln:append-score} in Algorithm~\ref{alg:calibrate})~\cite{chambers2018graphical}. Jitter helps reduce ties in the scores, ensuring continuity and smoother quantile transitions. To estimate the CDFs of \(\probD_\minor\) and \(\probD_\major\), the algorithm employs the \emph{plug-in method} based on observed data $D$~\cite{van2000asymptotic, wasserman2006all}. Specifically, it uses sorted score samples (\(\text{scores}_\minor\) and \(\text{scores}_\major\)) along with the position parameter \(\text{pos}\) to empirically estimate quantile values. Although alternatives such as kernel density estimation or parametric fitting exist, the plug-in method is favored here for its simplicity, robustness, and consistency, facilitating effective computation of the barycenter's quantile function.

The second point concerns the input dataset \(D = \{ \rPair{p}_1, \dots, \rPair{p}_{|D|} \}\), used to estimate \(\probD_\minor\) and \(\probD_\major\). This dataset does not require ground-truth labels--only matching scores are needed, which can be obtained by applying the score function \(s\).
As a result, \(s\) can be calibrated as a black box without access to the training data used to learn \(s\), such as in deep learning-based matchers. This significantly broadens the applicability of the calibration method. The only assumption is that the pairs in \(D\) are i.i.d.\ samples from the space of possible record pairs and that the dataset is large enough to reliably estimate the score distributions. 
The primary goal of the algorithm is to calibrate the scores for one or more query pairs. When the number of query pairs is large enough (as discussed in Section~\ref{sec:calib-th}), the dataset $D$ can simply be the set of query points itself.
We examine the effect of choosing a dataset $D$ with a distribution different from that of the query points in Section~\ref{sec:exp-dataseteffect}.
We now describe each step of Algorithm~\ref{alg:calibrate}.

Algorithm~\ref{alg:calibrate} takes the dataset $D$, a biased score function $s$ (i.e., $\fBias(\ScoreFunc, \PR) > 0$) and a query pair $\rPair{p}$, and returns a calibrated score $\hat{s}(\rPair{p})$. This calibrated score approximates the optimal fair score $s^*$, which solves \fProblem for the fairness metric $\fMetric = \PR$, removing \DP bias. The algorithm begins by initializing two empty lists to store scores separately for minority and majority groups (Line~\ref{ln:init-scores}).
For each record pair in the dataset $D$, it identifies the group based on the sensitive attribute (Line~\ref{ln:get-group}) and appends the computed score from $s$, with added Gaussian noise for smoothing, to the appropriate group's list (Line~\ref{ln:append-score}). This noise also ensures that $s(p_i) + \mathcal{N}(0, \sigma^2)$ forms i.i.d. and continuous samples of $s(X)$, a property we later use in Theorem~\ref{th:biasrisk}.
After processing all record pairs, the score lists for both groups are sorted in descending order (Line~\ref{ln:sort-scores}).
Note that as long as the observed dataset $D$ remains unchanged, these parts of the algorithm need to be executed only once for more than one query pair.

To calibrate the score for a query pair $\rPair{p}$, the algorithm identifies the group $g$ of $\rPair{p}$ (Line~\ref{ln:get-query-group}) and initializes a $pos$ variable within the sorted score list (Line~\ref{ln:init-pos}). 
It then iterates through the sorted scores to locate the position where $s(\rPair{p})$ would fit, counting scores greater than $s(\rPair{p})$ within group $g$ (Lines~\ref{ln:while-loop}–\ref{ln:increment-pos}). This position, relative to the list length, defines a quantile $q$, representing the percentage of scores in group $g$ greater than $s(\rPair{p})$ (Line~\ref{ln:compute-alpha1000}).
The algorithm then calculates $\text{pos}_{\bar{g}} = \left\lceil q\times |\text{scores}_{\bar{g}}| \right\rceil$, where $\bar{g}$ is the opposite group (i.e., if $g = a$, then $\bar{g} = b$, and vice versa). This position is used to find the score in the other group's list that corresponds to the same quantile $q$ as the query score in its own group.

Finally, the algorithm returns the calibrated score as a weighted average of the scores at the identified positions in both groups' sorted lists. 
The returned calibrated score is $\alpha \times \text{scores}_\minor[\text{pos}_\minor] + (1 - \alpha) \times \text{scores}_\major[\text{pos}_\major]$ (Line~\ref{ln:return-calibrated}).
Here, $\alpha$ is the proportion of minority group scores in $D$ (Line~\ref{ln:compute-alpha}).
Now, we provide an example to illustrate the algorithm's steps.

\begin{figure}[h]
    \centering
    \vspace{5mm}
    \includegraphics[width=0.8\linewidth]{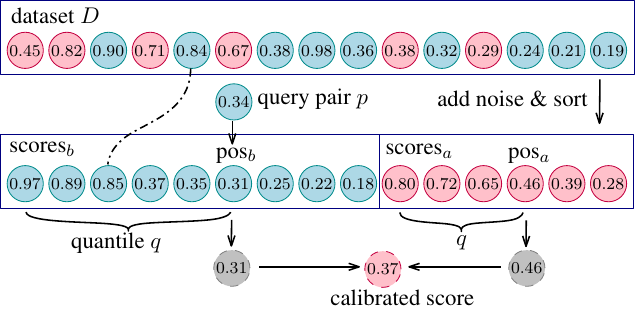}
    \caption{\normalsize Running Algorithm~\ref{alg:calibrate} in Example~\ref{ex:calibrate}}
    \label{fig:enter-label1}
\end{figure}

\begin{example} \label{ex:calibrate}\em
Consider dataset $D$ shown in Figure~\ref{fig:enter-label1}, containing 15 record pairs scored by $s$ and labeled \minor or \major by a sensitive attribute. The pairs and scores are: $(0.45, \minor)$, $(0.82, \minor)$, $(0.90, \major)$, $(0.71, \minor)$, $(0.84, \major)$, $(0.67, \minor)$, $(0.38, \major)$, $(0.98, \major)$, $(0.36, \major)$, $(0.38, \minor)$, $(0.32, \major)$, $(0.29, \minor)$, $(0.24, \major)$, $(0.21, \major)$, $(0.19, \major)$. Following the calibration algorithm, Gaussian noise $\mathcal{N}(0,0.05^2)$ is added to each score, resulting in the following (rounded) values: $(0.46, \minor)$, $(0.80, \minor)$, $(0.89, \major)$, $(0.72, \minor)$, $(0.85, \major)$, $(0.65, \minor)$, $(0.37, \major)$, $(0.97, \major)$, $(0.35, \major)$, $(0.39, \minor)$, $(0.31, \major)$, $(0.28, \minor)$, $(0.25, \major)$, $(0.22, \major)$, $(0.18, \major)$. Sorting these in descending order gives $\text{scores}_\major = [0.97, 0.89, 0.85, 0.37, 0.35, 0.31, 0.25, 0.22, 0.18]$ and $\text{scores}_\minor = [0.80, 0.72, 0.65, 0.46, 0.39, 0.28]$. Suppose the query pair $\rPair{p}$ has a score $s(\rPair{p}) = 0.34$ and belongs to the \major group. 
The algorithm sets $g = \major$ and $\text{pos} = 0$. It then performs a linear search to find the position of $0.34$ within $\text{scores}_\major$, which falls between $0.31$ and $0.35$, so $\text{pos}_\major = 6$. Next, $\alpha$ is calculated as the proportion of minority scores, $\alpha =|\text{scores}_\minor|/|D| = 6/15 = 0.4$, and the quantile $q = \text{pos}_{g}/|\text{scores}_g| = 6/9 \approx 0.67$. 
Using $q$, we compute the corresponding position in the minority scores: $\text{pos}_\minor = \left\lceil q \times |\text{scores}_{\bar{g}}| \right\rceil = 0.67 \times 6 \approx 4$, giving the score $0.46$ in the minority group. Finally, we compute a weighted average of the scores $0.31$ and $0.46$ using weights $\alpha$ and $1 - \alpha$, resulting in the calibrated score $\hat{s}(\rPair{p})$ = $0.4 \times 0.46$ + $0.6 \times 0.31$ = $0.37.$ Thus, the algorithm returns a calibrated score of $0.37$ for the query pair $\rPair{p}$. The steps are illustrated in Figure~\ref{fig:enter-label1}.
\boxtheorem\end{example}

\subsection{Theoretical Analysis of \calib} \label{sec:theo-calib}

Our analysis follows standard arguments from the optimal-transport and fair-regression literature for 1D distributional alignment and Wasserstein barycenters~\cite{villani2009optimal,peyre2019computational,chzhen2020fair}.
Algorithm~\ref{alg:calibrate} returns, for each input pair, a calibrated score and therefore defines a new score function $\hat{\ScoreFunc}$. In this section, $\ScoreFunc$ denotes the original score function, $\hat{\ScoreFunc}$ the calibrated score, and $\ScoreFunc^*$ the optimal fair score in the sense of Definition~\ref{df:fair-problem}. Let $n_\minor$ and $n_\major$ be the numbers of calibration samples in each group, and let $n=\min(n_\minor,n_\major)$.

We state mild assumptions used in our analysis in this section. Intuitively, these assumptions ensure that the score distributions behave ``smoothly,'' so that quantiles are uniquely defined and the empirical version of \calib\ is a stable approximation of its population counterpart. Technically, they allow us to apply standard empirical-process tools (e.g., Dvoretzky--Kiefer--Wolfowitz and quantile-process convergence) and guarantee that the Wasserstein barycenter map is well-defined and unique in one dimension. Importantly, these conditions are not restrictive in practice: modern matching or similarity models produce real-valued scores that can always be rescaled to $[0,1]$, and introducing an arbitrarily small jitter has no effect on application-level decisions—its role is purely to avoid pathological ties in the theoretical analysis.

The following are the assumptions. 
\begin{itemize}[leftmargin=3mm]
    \item {\em Assumption~1}: For each group $g\in\{\minor,\major\}$, the random variable $S_g$ (i.e., $\ScoreFunc(X)$ for $A=g$) has an absolutely continuous distribution on $[0,1]$ with CDF $F_g$ and quantile function $Q_g = F_g^{-1}$. The two distributions share the same support $[0,1]$, ensuring that both groups can be transported to a common barycenter in a well-defined way.
    \item {\em Assumption~2}: The calibration samples are i.i.d., and Algorithm~\ref{alg:calibrate} adds an independent Gaussian jitter to break ties, ensuring that empirical CDFs and quantiles are well-defined and converge uniformly to their population counterparts.
\end{itemize}

Under these assumptions, the population version of $\calib$ maps each group to the Wasserstein--2 barycenter of the two group-wise score distributions. Let $\alpha=\Pr(A=\minor)$ and define the barycenter quantile
\[
\tilde{Q}(p) = \alpha\, Q_\minor(p) + (1-\alpha)\, Q_\major(p), \qquad p\in[0,1].
\]
The corresponding \emph{population barycenter mapping} is
\[
\tilde{s}(x) = \tilde{Q}(F_{A(x)}(\ScoreFunc(x))),
\]
which ensures that $\tilde{s}(X)\mid A=\minor$ and $\tilde{s}(X)\mid A=\major$ have identical distributions. As standard in 1D optimal transport, $\tilde{s}$ uniquely minimizes the risk among all DP-fair score mappings and therefore coincides with $\ScoreFunc^*$.

Algorithm~\ref{alg:calibrate} computes the empirical analogue of this map by replacing each $F_g$ and $Q_g$ with their empirical counterparts. The following theorem shows that this empirical map converges to $\ScoreFunc^*$ at the parametric rate.

\begin{theorem}\label{th:biasrisk}
Let $\hat{\ScoreFunc}$ be the calibrated score produced by Algorithm~\ref{alg:calibrate} and let $\ScoreFunc^*$ denote the population barycenter score defined above. Under Assumptions~1--2,
\[
\fBias(\hat{\ScoreFunc}, \PR) = O_p(n^{-1/2}), 
\qquad 
\fRisk(\ScoreFunc^*, \hat{\ScoreFunc}) = O_p(n^{-1/2}),
\]
where $n=\min(n_\minor,n_\major)$.
\end{theorem}

\begin{proof} {\em Bias bound:} For any score function $s$, recall that
\[
\PR_g(s,\theta)=\Pr(s(X)\ge\theta\mid A=g)
= 1 - F_g^s(\theta),
\]
where $F_g^s$ is the CDF of $s(X)\mid A=g$. Hence
\[
\big|\PR_\minor(\hat{\ScoreFunc},\theta)-\PR_\major(\hat{\ScoreFunc},\theta)\big|
= \big|F_\minor^{\hat{s}}(\theta)-F_\major^{\hat{s}}(\theta)\big|.
\]
Thus
\[
\fBias(\hat{\ScoreFunc},\PR)
= \int_0^1 \big|F_\minor^{\hat{s}}(\theta)-F_\major^{\hat{s}}(\theta)\big|\, d\theta.
\]

In the population, both groups are mapped to the same barycenter distribution, so $F_\minor^{\tilde{s}}=F_\major^{\tilde{s}}$. With finite samples, Algorithm~\ref{alg:calibrate} replaces $F_g$ by the empirical CDF $\hat{F}_g$. By the Dvoretzky--Kiefer--Wolfowitz inequality,
\[
\sup_{\theta\in[0,1]} |\hat{F}_g(\theta)-F_g(\theta)| = O_p(n_g^{-1/2}),
\qquad g\in\{\minor,\major\}.
\]
Since the population barycenter CDFs coincide, the sup-norm difference between the two empirical calibrated CDFs is bounded by the sum of their uniform deviations:
\[
\sup_{\theta\in[0,1]} 
\big|F_\minor^{\hat{s}}(\theta)-F_\major^{\hat{s}}(\theta)\big|
= O_p(n^{-1/2}).
\]
Integrating over $\theta$ preserves this rate, giving
\[
\fBias(\hat{\ScoreFunc},\PR)=O_p(n^{-1/2}).
\]

\noindent {\em Risk bound:} The risk satisfies
\[
\fRisk(\ScoreFunc^*,\hat{\ScoreFunc})
= \E\!\left[\,|\ScoreFunc^*(X)-\hat{\ScoreFunc}(X)|\,\right].
\]
Since $\ScoreFunc^*=\tilde{s}$, it suffices to control the deviation between the empirical barycenter and the population barycenter in quantile space.

Let $Q_{g,n}$ be the empirical quantile of group $g$. The empirical barycenter quantile is
\[
\tilde{Q}_n(p) = \alpha\, Q_{\minor,n}(p) + (1-\alpha)\, Q_{\major,n}(p).
\]
Standard quantile-process theory gives
\[
\E\!\left[\,|Q_{g,n}(U)-Q_g(U)|\,\right] = O(n_g^{-1/2}),
\qquad g\in\{\minor,\major\},
\]
where $U\sim\mathrm{Unif}[0,1]$. Hence
\[
\E\!\left[\,|\tilde{Q}_n(U)-\tilde{Q}(U)|\,\right] = O_p(n^{-1/2}).
\]
Mapping back through $F_{A(x)}$ preserves the order, so
\[
\fRisk(\ScoreFunc^*, \hat{\ScoreFunc}) = O_p(n^{-1/2}).
\]

\medskip
Thus both DP bias and risk converge at the parametric rate $O_p(n^{-1/2})$. We use $O_p(\cdot)$ because the bounds involve random empirical CDFs and quantiles whose fluctuations shrink at rate $n^{-1/2}$ only in probability.
\end{proof}

\medskip
Theorem~\ref{th:biasrisk} shows that as $n$ grows, the calibrated score $\hat{\ScoreFunc}$ approaches the optimal DP-fair score $\ScoreFunc^*$, simultaneously achieving fairness and minimizing deviation from the original scores.

\section{Conditional Score Calibration}\label{sec:partial}

Although the \calib algorithm ensures \DP, it does not satisfy other fairness criteria, such as \EOD and \EO, because these definitions rely on the true (matching) labels of pairs- a factor not considered by \calib. To address this, we introduce \emph{conditional calibration} (\ccalib), as shown in Algorithm~\ref{alg:pcalibrate}. \ccalib estimates the unknown true labels, enabling it to meet fairness definitions that depend on label information. Like the original calibration algorithm, \ccalib reduces the correlation between $\plabel$ and $\senAtt$, but it conditions this adjustment on $\tlabel$, enforcing $\plabel \indep \senAtt \mid \tlabel$. This produces a fair score distribution that meets fairness criteria such as \EO and \EOD.

\begin{algorithm}
\caption{$\ccalib(D,s,\rPair{p})$}\label{alg:pcalibrate}
\KwInput{$D = \{\rPair{p}_1,...,\rPair{p}_{|D|}\}$, $s$, and a query pair $\rPair{p}$}
\KwOutput{Calibrated score for $\rPair{p}$}
    $\text{scores}_\major \leftarrow []$; $\text{scores}_\minor \leftarrow []$; \label{ln-p:init-scores}\\     
    $\cthreshold\leftarrow \text{meanshift} (D)$; \label{ln:gammaestimate} \algCmnd{Estimate decision threshold}\\
    $\text{match}\leftarrow \mathbbm{1}(s(\rPair{p}) \ge \cthreshold)$; \algCmnd{Label query pair}\label{ln:cthredhold1}\\
    $\text{size}\leftarrow 0$;\hfill \algCmnd{Size of matched pairs set}\\
    
    \ForEach{$\rPair{p}_i \in D$}{\label{ln-p:for-each}
    \Indp $g \leftarrow \rPair{p}_i[\senAtt]$;\label{ln-p:get-group}\\
    $\text{match}_i\leftarrow \mathbbm{1}(s(\rPair{p}_i) \ge \cthreshold)$; \hspace{-2mm}\algCmnd{Labels pairs}\label{ln:cthredhold2}\\
    \If{$\textit{match}$ = $\text{match}_i$\label{ln-p:if}}{
    $\text{append}(\text{scores}_g, s(\rPair{p}_i) + \mathcal{N}(0, \sigma^2))$;\\
    $\text{size}\leftarrow \text{size}+1$;
    \label{ln-p:append-score}\\
    \Indm }}
    $\text{sort}(\text{scores}_\minor)$; 
    $\text{sort}(\text{scores}_\major)$;
    $g \leftarrow \rPair{p}[\senAtt]$; $\text{pos}_g \leftarrow 0$; \label{ln-p:init-pos}\\
    \While{$\text{pos}_g < |\text{scores}_g|$ \textbf{\em and} $\text{scores}_g[\text{pos}_g] > s(\rPair{p})$}{\label{ln-p:while-loop}
    \Indp $\text{pos}_g \leftarrow \text{pos}_g + 1$; \label{ln-p:increment-pos}\\
    \Indm }
    $\alpha \leftarrow \frac{|\text{scores}_\minor|}{\text{size}}$; $q\leftarrow \frac{\text{pos}_{g}}{|\text{scores}_g|}$;\label{ln-p:compute-alpha-end}\\ %\label{ln-p:compute-alpha}
    $\text{pos}_{\bar{g}} \leftarrow \left\lceil q\times |\text{scores}_{\bar{g}}| \right\rceil$; \\ %\label{ln-p:compute-alpha}
    \Return{$\alpha \times \text{scores}_\minor[\text{pos}_\minor] + (1-\alpha) \times \text{scores}_\major[\text{pos}_\major]$; \label{ln-p:return-calibrated}}
\end{algorithm}

% meanshift is a non-parametric clustering algorithm that identifies clusters in a dataset by iteratively shifting data points toward regions of higher density. When working with a set of scores and no ground truth labels, meanshift can be employed to automatically detect the natural grouping of the data. In this context, the algorithm helps determine an optimal threshold by finding two clusters within the score distribution. The threshold is then set as the midpoint between the centers of these two clusters, effectively separating the data into two classes without the need for labeled training examples.

\subsection{\ccalib Algorithm}

The core idea behind Algorithm~\ref{alg:pcalibrate} is to incorporate predicted labels using variables $\text{match}$ and $\text{match}_i$. Here, $\text{match}$ is the predicted label for the query pair $p$, while $\text{match}_i$ is the predicted label for each pair in $D$. This is illustrated in Lines~\ref{ln:cthredhold1} and \ref{ln:cthredhold2} of Algorithm~\ref{alg:pcalibrate}. To estimate labels---whether for the query pair or pairs in $D$---we require a decision threshold to assign predicted labels based on scores. In traditional binary classification, this threshold is determined on a labeled validation set to optimize a performance metric (e.g., F1-score) and then applied to unlabeled test data. In our case, we only have a query pair and a set $D$ of observed pairs. To choose an effective decision threshold, we use a non-parametric clustering algorithm, \emph{meanshift}~\cite{comaniciu2002mean}, as shown in Line~\ref{ln:gammaestimate}. The meanshift algorithm places a window around each data point, computes the mean within the window, and shifts the window toward the mean until convergence. This process clusters points around the modes of the distribution. When applied to score values, which tend to cluster near 0 and 1 (which is the case for record matching problem), the algorithm identifies two clusters. The decision threshold, denoted as $\cthreshold$, is then set as the midpoint between the centers of these clusters, effectively separating the data into two classes without the need for labeled data.

The parameter $\cthreshold$ has two purposes in \ccalib. In Line~\ref{ln:cthredhold1}, it estimates the label of query pair, and in Line~\ref{ln:cthredhold2}, it separates the samples in $D$ into positive and negative samples, assuming $D$ is unlabeled. 
Unlike decision threshold of $\threshold$ in the final down-stream applications, $\cthreshold$ is specific to the calibration algorithm and does not decide the final label of $\rPair{p}$.\ignore{, though it implicitly treats it as positive or negative in Line~\ref{ln:cthredhold1}. If $D$ contains labeled samples, $\cthreshold$ is not required in Line~\ref{ln:cthredhold1}, making the calibration method applicable even when $D$ is unlabeled.} While $\cthreshold$ might appear inconsistent with a threshold-independent algorithm, it acts as a tuning parameter rather than a classification threshold. By ``threshold-independent,'' we mean the algorithm mitigates bias across all classification thresholds in the final results.

Moving forward, using the conditional check in Line~\ref{ln-p:if}, the algorithm adjusts scores based on the query pair’s predicted label. If the query pair is predicted as positive, only pairs in $D$ predicted as positive are included in the calibration; the same applies for a negative prediction. The remainder of the algorithm follows the process of Algorithm~\ref{alg:calibrate} described in Section~\ref{sec:calib}. We call this approach ``conditional'' because it calibrates only among pairs sharing the same predicted label as the query pair, consequently enforcing \EOD and \EO.
Now, we demonstrate the entire process with the following example.

% \mhm{This algorithm uses a parameter $\cthreshold$, along with the scores, to estimate the true labels. Based on these estimated labels, the calibration process is applied. The parameter $\cthreshold$ is critical in this algorithm and is selected using validation data to determine the value that minimizes the target bias metric. For effective tuning, the test and validation data must share the same distribution. Tuning $\cthreshold$ using validation data ensures that calibration reduces bias consistently across all thresholds in the test data. }

\begin{figure}
    \centering
    \includegraphics[width=0.77\linewidth]{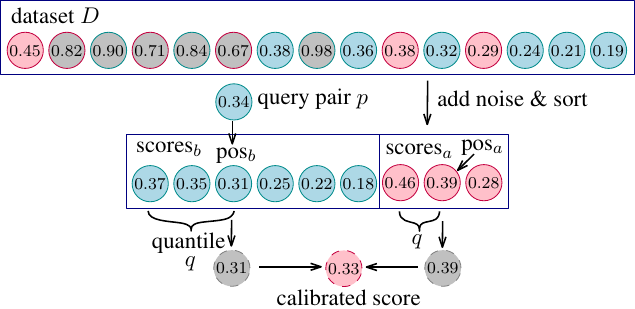}
    \caption{Running Algorithm~\ref{alg:pcalibrate} in Example~\ref{ex:pcalibrate}}
    \label{fig:enter-label2}
\end{figure}

\begin{example} \label{ex:pcalibrate}\em
Following Example~\ref{ex:calibrate}, consider a dataset $D$ containing 15 pairs with scores as shown in Figure~\ref{fig:enter-label2}. Using meanshift, the decision threshold is set to $\cthreshold=0.57$, and the query pair $\rPair{p}$ is predicted as 0 (nonmatch). 
As a result, the calibration process is applied only to pairs predicted as nonmatches (label 0). As shown in Figure~\ref{fig:enter-label2}, the matched pairs are filtered out, and the remaining pairs are calibrated as in Algorithm~\ref{alg:calibrate}. This leads to sorted scores in descending order $\text{scores}_\major = [0.37, 0.35, 0.31, 0.25, 0.22, 0.18]$ and $\text{scores}_\minor = [0.46, 0.39, 0.28]$. Next, we calculate $\text{size} = 9$ (the total number of non-matching pairs in $D$ to be used instead of $|D|$ for computing $\alpha$), with $|\text{scores}_\minor| = 3$ and $|\text{scores}_\major| = 6$. The positions are $\text{pos}_\minor=2$ and $\text{pos}_\major=3$ with quantile $q=0.5$. We also compute \(\alpha = |\text{scores}_\minor|/\text{size} = 3/9 = 0.33\). The calibrated score is calculated as $\hat{s}(\rPair{p})$ = $0.33 \times 0.39$ + $0.67 \times 0.31\approx 0.33.$ Therefore, the CondCalibrate algorithm returns a calibrated score of $0.33$ for the query pair $\rPair{p}$.\boxtheorem
\end{example}

Proving the convergence of the calibrated score function $\hat{s}$ from \ccalib is challenging and depends on the quality of the initial score function, as it affects the estimation of labels for $D$ and the query pair $\rPair{p}$. Empirically, we show that the algorithm effectively reduces bias while maintaining low risk.

\subsection{Theoretical Analysis of \ccalib}\label{sec:calib-th}

Before presenting the theoretical guarantees for \ccalib, we introduce several quantities that describe the interaction between score calibration and label stratification. Let
\[
\varepsilon^\star \;=\; \inf_{\theta\in[0,1]} \Pr\!\bigl(Y \neq \mathbf{1}\{s(X)\ge\theta\}\bigr)
\]
be the \emph{irreducible thresholding error} of the original score $s$. When the class-conditional score distributions overlap, no threshold perfectly separates $Y{=}1$ from $Y{=}0$, and thus $\varepsilon^\star>0$; when they are separable, $\varepsilon^\star=0$. 

Relatedly, several works study how fairness guarantees degrade when protected information is only available through noisy measurements or proxy signals (e.g., noisy protected attributes)~\cite{celis2020noisy,mehrotra2022fair}. These results are complementary to our setting: they typically do not analyze OT/Wasserstein-barycenter score calibration, whereas our focus is on how conditioning distributional alignment on an estimated label $\hat Y$ affects \EO/\EOD.

\ccalib uses an unsupervised threshold $\cthreshold$ (e.g., estimated by mean-shift) to form predicted labels $\hat Y=\mathbf{1}\{s(X)\ge\cthreshold\}$. The realized mislabeling rate
\[
\varepsilon_n \;=\; \Pr(\hat Y\neq Y)
\]
depends on the sample size $n$ through the estimation of $\cthreshold$. It is convenient to write
\[
\varepsilon_n
\;=\;
\varepsilon^\star \,+\, (\varepsilon_n-\varepsilon^\star),
\]
where the excess term $(\varepsilon_n-\varepsilon^\star)$ captures the \emph{estimation error} introduced by learning the threshold. Under standard regularity (unique split point, smooth and nonvanishing densities), this term typically shrinks at the parametric rate $O_p(n^{-1/2})$.

We refer to $\{Y=1\}$ and $\{Y=0\}$ as the \emph{true-label strata} (positive and negative), and to $\{\hat Y=1\}$ and $\{\hat Y=0\}$ as the \emph{predicted-label strata}. \ccalib applies \calib separately within each predicted stratum: the predicted positive stratum drives the TPR component and the predicted negative stratum drives the FPR component. Consequently, the resulting Equalized Odds behavior depends on two sources of error:
(i) finite-sample estimation within each stratum, contributing an $O_p(n^{-1/2})$ term, and 
(ii) the quality of the stratification, governed by $\varepsilon^\star$ and $(\varepsilon_n-\varepsilon^\star)$.

\begin{theorem}\label{th:ccalib_eod}
Let $\hat{\ScoreFunc}$ be the score function produced by \ccalib, which (i) estimates a threshold $\cthreshold$ and labels $\hat Y=\mathbf{1}\{s(X)\ge\cthreshold\}$, and (ii) applies \calib within the predicted strata $\{\hat Y=1\}$ and $\{\hat Y=0\}$. Let %
\begin{align*}
n&=\min_{y\in\{0,1\},\,g\in\{\minor,\major\}} n_{y,g}\\
\varepsilon_n&=\Pr(\hat Y\neq Y)\\
\varepsilon^\star&=\inf_{\theta\in[0,1]} \Pr\!\bigl(Y \neq \mathbf{1}\{s(X)\ge\theta\}\bigr).
\end{align*}

\noindent Under the same regularity assumptions as for \calib, 
\begin{align*}
\fBias(\hat{\ScoreFunc},\TPR)+\fBias(\hat{\ScoreFunc},\FPR)
= O_p(n^{-1/2}) + O\!\bigl(\varepsilon^\star + |\varepsilon_n-\varepsilon^\star|\bigr)
\\
\fRisk(s^\star,\hat{\ScoreFunc})
= O_p(n^{-1/2}) + O\!\bigl(\varepsilon^\star + |\varepsilon_n-\varepsilon^\star|\bigr),
\end{align*}
where $s^\star$ is an oracle EOD-fair map. In particular, when $\varepsilon^\star=0$ (threshold-separable classes) and $\cthreshold$ is consistent (so $|\varepsilon_n-\varepsilon^\star|=O_p(n^{-1/2})$), both the EOD bias and the risk converge at the parametric rate $O_p(n^{-1/2})$.
\end{theorem}

\begin{proof}[Proof sketch]
If we could stratify by the true label $Y$, applying \calib\ within $\{Y=1\}$ and $\{Y=0\}$ would align the group-wise CDFs in each stratum. As in Theorem~\ref{th:biasrisk}, empirical-process fluctuations of CDFs and quantiles contribute an $O_p(n^{-1/2})$ error to both TPR and FPR components, and similarly to the risk.

\ccalib\ instead stratifies using $\hat Y$, so each predicted stratum is a mixture of the two true strata. The mixing proportions are governed by the mislabeling rate $\varepsilon_n$, yielding an $O(\varepsilon_n)$ perturbation of the stratum-wise CDFs. By Lipschitz continuity of quantiles in one dimension, these perturbations propagate linearly through the barycenter map computed by \calib, producing an additional $O(\varepsilon_n)$ contribution to both EOD components (the \TPR and \FPR terms) and to the risk.

Finally, writing $\varepsilon_n=\varepsilon^\star+(\varepsilon_n-\varepsilon^\star)$ separates irreducible overlap and threshold-estimation error, completing the bound.
\end{proof}

\section{Experiments}
\label{sec:experiments}

% In our experiments, first we assess bias in existing matching methods using our score bias metric. Second, we show that this metric captures bias in matching tasks better than traditional binary measures (Section~\ref{sec:exp-bias}). Third, we evaluate our calibration algorithms, \calib (Algorithm~\ref{alg:calibrate}) and \ccalib (Algorithm~\ref{alg:pcalibrate}), for reducing score bias while keeping risk low. Bias reduction is discussed in Sections~\ref{sec:exp-calibrate} and~\ref{sec:exp-con-calibrate}, with risk analyzed in Section~\ref{sec:exp-risk}. Finally, we examine the effect of different input dataset $D$ selction in Secion ~\ref{sec:exp-dataseteffect}.

In our experiments, we first assess bias in existing matching methods using our score bias metric. Second, we show that this metric captures bias in matching tasks more effectively than traditional binary measures (Section~\ref{sec:exp-bias}). Third, we evaluate our calibration algorithms for their ability to reduce score bias while maintaining low risk. Bias reduction results are presented in Sections~\ref{sec:exp-calibrate} and~\ref{sec:exp-con-calibrate}, with risk analysis in Section~\ref{sec:exp-risk}. Finally, we examine the impact of different input dataset $D$ selections in Section~\ref{sec:exp-dataseteffect}.

% Our experiments have three main objectives. First, we evaluate the bias in existing matching methods using our concept of matching score bias. Second, we demonstrate that matching score bias better captures bias in matching tasks compared to existing binary bias notions. Both objectives are addressed in Section~\ref{sec:exp-bias}. Third, we assess the effectiveness of our calibration algorithms, \calib (Algorithm~\ref{alg:calibrate}) and \ccalib (Algorithm~\ref{alg:pcalibrate}), in reducing score biases while keeping risk low. Bias reduction is analyzed in Sections~\ref{sec:exp-calibrate} and~\ref{sec:exp-con-calibrate}, and risk is studied in Section~\ref{sec:exp-risk}.

\subsection{Experimental Setup}  \label{sec:setup}

Before presenting our results, we describe the experimental setup, including datasets and matching methods. All implementations and findings are available in our repository~\cite{moslemi2024fairEM}.

\subsubsection{Datasets} \label{sec:datasets}
% We use datasets from well-established benchmarks: Beer (\Beer), Walmart-Amazon ($\walamz$), Amazon-Google ($\amzgog$), Fodors-Zagat (\FodorZag), iTunes-Amazon ($\itunamz$), DBLP-GoogleScholar (\DBLPGogS), and DBLP-ACM (\DBLPACM). These datasets are widely used in research to evaluate record matching systems~\cite{mudgal2018deep}. Key statistics are summarized in Table~\ref{tab:datasets}.

% These datasets do not have a distinct column labeled as a sensitive attribute. Instead, we identify minority and majority groups based on specific criteria applied to certain columns, as described in previous research~\cite{nilforoushan2022entity,shahbazi2023through,moslemi2024threshold}. The minority pairs are defined by the following entities:

We used record matching benchmarks~\cite{mudgal2018deep}: Beer (\Beer), Walmart-Amazon (\walamz), Amazon-Google (\amzgog), Fodors-Zagat (\FodorZag), iTunes-Amazon (\itunamz), DBLP-GoogleScholar (\DBLPGogS), and DBLP-ACM (\DBLPACM). Table~\ref{tab:datasets} summarizes statistics for each dataset. None of these datasets explicitly denotes a sensitive attribute.

For \DBLPACM, we use a gender proxy (female first names appearing in the authors field) to form groups, which serves as an example of a sensitive attribute commonly studied in fairness. For the remaining benchmarks, the group definitions (e.g., genre, manufacturer, keyword-based slices) should be viewed as benchmark-derived subpopulations rather than claims about inherently sensitive attributes. Our goal in using these established slices, following prior work~\cite{nilforoushan2022entity,shahbazi2023through,moslemi2024threshold}, is to (i) evaluate whether calibration can repair matching scores whenever systematic score disparities are observed between groups, and (ii) enable direct comparison with existing record-matching methods and fairness-aware entity matching frameworks that use the same benchmarks and group constructions.

We follow prior work~\cite{nilforoushan2022entity,shahbazi2023through,moslemi2024threshold} to define minority and majority based on the following criteria:

\begin{itemize}[leftmargin=0.35cm]
    \item \DBLPACM: the ``authors'' attribute contains a female name.
    \item \FodorZag: the ``Type'' attribute equal to ``Asian.''
    \item \amzgog: with ``Microsoft'' as the ``manufacturer.''
    \item \walamz: the ``category'' attribute equal to ``printers.''
    \item \DBLPGogS: ``venue'' containing ``vldb j.''
    \item \Beer: ``Beer Name'' contains ``red.''
    \item \itunamz: the ``Genre'' attribute containing ``Dance.''
\end{itemize}

\begin{table}[ht!]
    \centering
    \normalsize
    \setlength{\tabcolsep}{3mm} % Adjust global column spacing
    \resizebox{\textwidth}{!}{
        \begin{tabular}{l c c c c c l}
    \toprule
    \textbf{Dataset} & \textbf{\#Attr.} & \multicolumn{2}{c}{\textbf{Dataset Size}} & \multicolumn{2}{c}{\textbf{Eq\%}} & \textbf{Sens. Attr.} \\
    \cmidrule(lr){3-4} \cmidrule(lr){5-6}
    & & \textbf{Train} & \textbf{Test} & \textbf{Train} & \textbf{Test} & \\
    \midrule
    \walamz   & 5 & 8,193 & 2,049 & 9.39 & 9.42 & Category     \\ 
    \Beer     & 4 & 359 & 91 & 15.04 & 15.38 & Beer Name    \\ 
    \amzgog   & 3 & 9,167 & 2,293 & 10.18 & 10.21 & Manufacturer \\ 
    \FodorZag & 6 & 757 & 190 & 11.62 & 11.58 & Type         \\ 
    \itunamz  & 8 & 430 & 109 & 24.42 & 24.77 & Genre        \\ 
    \DBLPGogS & 4 & 22,965 & 5,742 & 18.62 & 18.63 & Venue        \\ 
    \DBLPACM  & 4 & 9,890 & 2,473 & 17.96 & 17.95 & Authors      \\ 
    \bottomrule
    \end{tabular}
}
\caption{Datasets and their characteristics. {\em Dataset Size} shows training and test set sizes. {\em Eq\%} columns represent the percentage of equivalence pairs in the training and test data.}
    \label{tab:datasets}
\end{table}

\subsubsection{Record Matching Methods} \label{sec:methods}
We selected five high-performing deep learning record-matching methods with diverse architectures to demonstrate our calibration method's versatility.
\dmatcher combines tokenized inputs, pre-trained embeddings, and metadata, achieving strong performance on structured data~\cite{mudgal2018deep}.
 \hiergat employs a Hierarchical Graph Attention Transformer for contextual embeddings~\cite{yao2022entity}. \ditto enhances pre-trained language models with text summarization, data augmentation, and domain expertise~\cite{li2020deep}. \EMT uses transformers to model relationships across attributes~\cite{brunner2020entity}. \hmatcher builds on \dmatcher with cross-attribute token alignment and attribute-aware attention~\cite{fu2021hierarchical}.
 Each model is trained for 100 epochs, with early stopping if the F1 score fails to improve for 15 epochs. The data are split 75\% for training and 25\% for validation. After training, the model generates scores for the test set, which feed into our calibration algorithms.

\subsection{Experimental Results}\label{sec:results}

We now present our results. For simplicity, we denote matching score biases $s$ as $\DP$, $\EO$, and \EOD, corresponding to $\fBias(s,\PR)$, $\fBias(s,\TPR)$, and $\fBias(s,\TPR)+\fBias(s,\FPR)$, respectively. We use $\DP_\theta$, $\EO_\theta$, and $\EOD_\theta$ to represent the corresponding threshold-specific (binary) fairness metrics for a matcher with score function $s$ and threshold $\theta$ (e.g., $\DP_{0.5}$ for $\theta=0.5$).

\subsubsection{Biases in Matching Scores}\label{sec:exp-bias}

Here, we analyze the matching methods and their associated biases. Table~\ref{tab:bias} summarizes these biases across datasets and matching models. All bias values are reported as percentages (\%). Each metric is reported for three different thresholds, alongside the bias metric for score values as defined in this work. Due to space limitations, we present a subset of model-dataset combinations here; the complete results are available in our repository~\cite{moslemi2024fairEM}. An initial look at the table shows that traditional fairness metrics can vary significantly across thresholds. For example, the biases highlighted in blue indicate a notable disparity with $\EO_{0.1}=14\%$, but much lower biases for $\EO_{0.5}=1.35\%$ and $\EO_{0.95}=3.38\%$. Similarly, the biases highlighted in green show minimal bias for $\DP_{0.1}=0.05\%$ but much higher biases for $\DP_{0.5}=4.85\%$ and $\DP_{0.95}=3.49\%$. Similar trends are evident for \EOD across methods and datasets. The score bias metrics shown with $\DP$, $\EO$, and \EOD provide an aggregate view of biases across all thresholds, capturing overall bias effectively.
The results in the table also reveal differences in biases across methods and datasets. In most cases, substantial biases are present, indicating a clear need for debiasing methods.

\begin{table}[ht!]
    \centering
    \renewcommand{\arraystretch}{1} % Adjust row height
    \setlength{\tabcolsep}{1.5mm} % Adjust global column spacing
    % \resizebox{\textwidth}{!}{
    \normalsize

\begin{tabular}{l@{\hskip 1mm}l@{\hskip 2mm}c@{\hskip 2mm}c@{\hskip 2mm}c@{\hskip 2mm}c@{\hskip 2mm}c@{\hskip 2mm}c@{\hskip 2mm}c@{\hskip 2mm}c@{\hskip 2mm}c@{\hskip 0.3mm}c@{\hskip 2mm}c@{\hskip 2mm}c}
\toprule
 & &  & \multicolumn{3}{c}{\footnotesize $\DP_\theta$} &  & \multicolumn{3}{c}{\footnotesize $\EO_\theta$} &   & \multicolumn{3}{c}{\footnotesize $\EOD_\theta$} \\
\cmidrule(lr){4-6} \cmidrule(lr){8-10}  \cmidrule(lr){12-14} 

& \footnotesize Dataset & \footnotesize $\DP$ & \footnotesize $0.1$ & \footnotesize $0.5$ & \footnotesize $0.95$ & \footnotesize $\EO$ & \footnotesize $0.1$ & \footnotesize $0.5$ & \footnotesize $0.95$ & \footnotesize $\EOD$ & \footnotesize $0.1$ & \footnotesize $0.5$ & \footnotesize $0.95$ \\
    \midrule

\multirow{4}{*}{\rotatebox[origin=c]{90}{ \footnotesize \dmatcher}}
& \footnotesize \FodorZag & \multicolumn{1}{!{\vrule width 1pt}c}{ 2.86} & 5.0 & 2.45 & 1.92 &  \multicolumn{1}{!{\vrule width 1pt}c}{ 5.5} & 0 & 5.3 & 15.8 &   \multicolumn{1}{!{\vrule width 1pt}c}{ 5.8} & 2.19 & 5.3 & 15.8 \\
& \footnotesize \DBLPGogS & \multicolumn{1}{!{\vrule width 1pt}c}{ 6.2} & 6.6 & 6.6 & 1.80 &  \multicolumn{1}{!{\vrule width 1pt}c}{ 11.5} & 7.6 & 10.8 & 7.9 &   \multicolumn{1}{!{\vrule width 1pt}c}{ 11.8} & 7.8 & 11.0 & 8.0 \\
& \footnotesize \amzgog & \multicolumn{1}{!{\vrule width 1pt}c}{ 8.9} & 13.5 & 9.6 & 2.30 &  \multicolumn{1}{!{\vrule width 1pt}c}{ {\cellcolor{SkyBlue!30}4.40}} & {\cellcolor{SkyBlue!30}14.0} & {\cellcolor{SkyBlue!30}1.35} & {\cellcolor{SkyBlue!30}3.38} &   \multicolumn{1}{!{\vrule width 1pt}c}{ 8.2} & 20.7 & 5.3 & 3.86 \\
& \footnotesize \DBLPACM & \multicolumn{1}{!{\vrule width 1pt}c}{ 3.60} & 4.50 & 3.90 & 0.82 &  \multicolumn{1}{!{\vrule width 1pt}c}{ 1.55} & 1.05 & 1.57 & 7.8 &   \multicolumn{1}{!{\vrule width 1pt}c}{ 1.80} & 1.51 & 1.70 & 7.8 \\
\midrule

\multirow{4}{*}{\rotatebox[origin=c]{90}{ \footnotesize \hiergat}}
& \footnotesize \FodorZag & \multicolumn{1}{!{\vrule width 1pt}c}{ 2.33} & 2.45 & 2.45 & 1.17 &  \multicolumn{1}{!{\vrule width 1pt}c}{ 7.1} & 5.3 & 5.3 & 15.8 &   \multicolumn{1}{!{\vrule width 1pt}c}{ 7.1} & 5.3 & 5.3 & 15.8 \\
& \footnotesize \DBLPGogS & \multicolumn{1}{!{\vrule width 1pt}c}{ 5.3} & 5.3 & 5.4 & 5.0 &  \multicolumn{1}{!{\vrule width 1pt}c}{ 3.40} & 3.83 & 2.85 & 3.95 &   \multicolumn{1}{!{\vrule width 1pt}c}{ 3.90} & 4.60 & 3.22 & 4.60 \\
& \footnotesize \amzgog & \multicolumn{1}{!{\vrule width 1pt}c}{ 11.1} & 14.8 & 11.7 & 4.14 &  \multicolumn{1}{!{\vrule width 1pt}c}{ 15.0} & 11.3 & 13.5 & 14.9 &   \multicolumn{1}{!{\vrule width 1pt}c}{ 19.5} & 18.3 & 18.2 & 15.8 \\
& \footnotesize \DBLPACM & \multicolumn{1}{!{\vrule width 1pt}c}{ 4.40} & 4.70 & 4.30 & 4.09 &  \multicolumn{1}{!{\vrule width 1pt}c}{ 0.44} & 0.80 & 0.27 & 0.01 &   \multicolumn{1}{!{\vrule width 1pt}c}{ 0.66} & 1.20 & 0.30 & 0.22 \\
\midrule

\multirow{4}{*}{\rotatebox[origin=c]{90}{ \footnotesize \hmatcher}}
& \footnotesize \FodorZag & \multicolumn{1}{!{\vrule width 1pt}c}{ 2.93} & 1.34 & 3.09 & 3.09 &  \multicolumn{1}{!{\vrule width 1pt}c}{ 0.50} & 0 & 0  & 0 &   \multicolumn{1}{!{\vrule width 1pt}c}{ 1.41} & 0.71 & 0 & 0 \\
& \footnotesize \DBLPGogS & \multicolumn{1}{!{\vrule width 1pt}c}{ 5.4} & 7.9 & 5.8 & 3.83 &  \multicolumn{1}{!{\vrule width 1pt}c}{ 4.40} & 1.15 & 5.6  & 8.9 &   \multicolumn{1}{!{\vrule width 1pt}c}{ 5.3} & 6.3 & 5.8 & 8.9 \\
& \footnotesize \amzgog & \multicolumn{1}{!{\vrule width 1pt}c}{ 9.1} & 1.84 & 10.7 & 4.28 &  \multicolumn{1}{!{\vrule width 1pt}c}{ 22.0} & 1.80 & 33.8  & 8.8 &   \multicolumn{1}{!{\vrule width 1pt}c}{ 26.3} & 5.1 & 38.1 & 9.9 \\
& \footnotesize \DBLPACM & \multicolumn{1}{!{\vrule width 1pt}c}{ {\cellcolor{Teal!30}3.89}} &  {\cellcolor{Teal!30}0.05} &  {\cellcolor{Teal!30}4.85}  & {\cellcolor{Teal!30}3.49} &  \multicolumn{1}{!{\vrule width 1pt}c}{ 1.04} & 0 & 0.79 & 0.64 &   \multicolumn{1}{!{\vrule width 1pt}c}{ 1.70} & 0.06 & 1.61  & 0.76 \\

    \bottomrule
    \end{tabular}
    % }

\caption{We evaluated the distributional disparity across models and datasets. Our metrics were compared against traditional fairness measures at thresholds of 0.1, 0.5, and 0.95. All values are reported in \%.}
            \label{tab:bias}
        
\end{table}

\begin{figure}[ht]
         \centering
        \subfloat[$\DP = 3.9\%$ ]{\label{fig:DPcompare}
			{\includegraphics[width=0.45\textwidth]{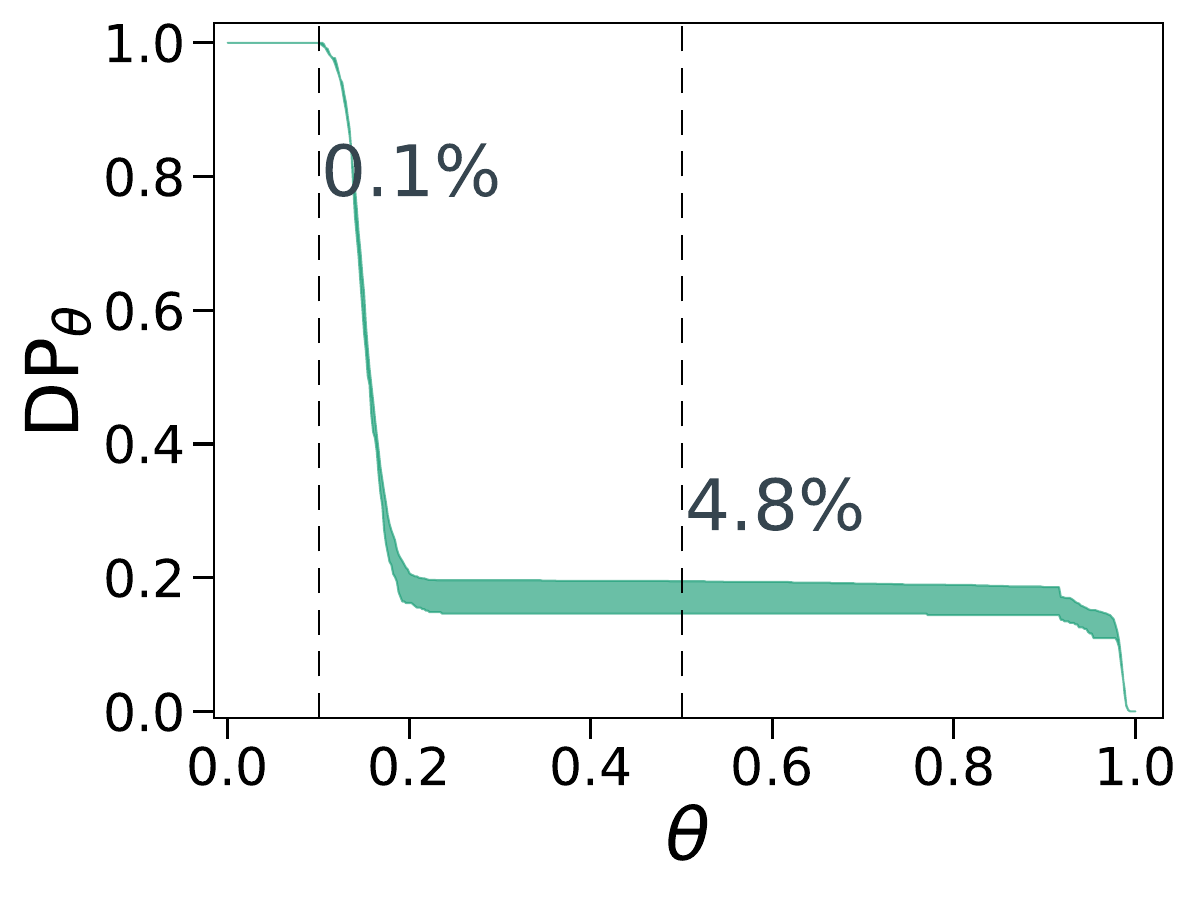}}}
        \hfill
        \subfloat[$\EO = 4.4\%$]{\label{fig:EOcompare}
			{\includegraphics[width=0.45\textwidth]{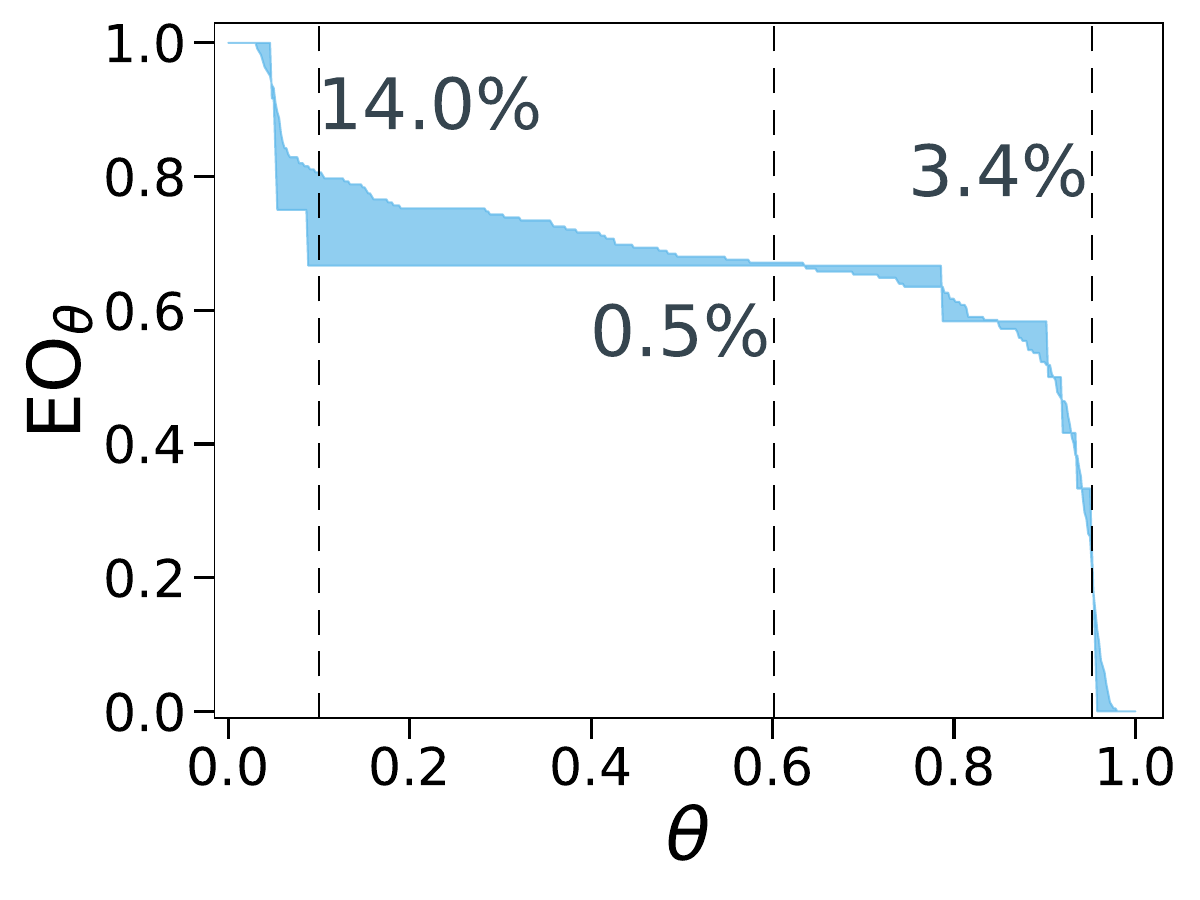}}}	
	\caption{Variation of bias across thresholds, highlighting the limits of single-threshold fairness assessments. Figure~\ref{fig:DPcompare} shows this for \DBLPACM dataset with \hmatcher, while Figure~\ref{fig:EOcompare} does for \amzgog dataset with \dmatcher.}
 \label{fig:compared}
\end{figure}

To further demonstrate the importance of biases for scores (Distributional \DP or \EO), we analyze two examples from Table~\ref{tab:bias} in more detail. These examples are highlighted in Table~\ref{tab:bias} and shown in Figure~\ref{fig:compared}. Figure~\ref{fig:DPcompare} illustrates $\DP_\theta$ across varying thresholds for \hmatcher on \DBLPACM, while Figure~\ref{fig:EOcompare} shows $\EO_\theta$ for \dmatcher on \amzgog. As seen in these figures, a model may appear fair at certain thresholds (with narrow color shade widths) based on traditional fairness metrics. However, at thresholds very close to these fair points, the model often displays significant bias toward one group, complicating fairness evaluation. In both figures, the entire colored area effectively represents the distributional disparity.

\subsubsection{Bias Reduction through Calibration}
\label{sec:exp-calibrate}

Here, we analyze the performance of \calib, which requires a set of unlabeled record pairs to estimate score distributions per group. For the input dataset $D$, we use the test inputs themselves (i.e., the set of query pairs) to estimate group-wise score distributions and calibrate their scores. Importantly, this step does \emph{not} use ground-truth match labels; it only uses the score function outputs on the given inputs. This choice aligns with the intended use case of post-processing a black-box matcher when only the deployment inputs are available.

This experimental protocol follows the standard i.i.d. assumption in machine learning, where training and test inputs are drawn from the same (or sufficiently similar) distribution. When there is substantial distribution shift between the data used to estimate score distributions (i.e., $D$) and the query inputs being calibrated, calibration quality can degrade; studying and addressing such distribution shift is beyond the scope of this work. We partially illustrate this sensitivity by comparing different choices of $D$ in Section~\ref{sec:exp-dataseteffect}.

Figure~\ref{fig:comparedDSPdp} presents selected results, showing a notable reduction in score bias $\DP$ before and after calibration across models and datasets. This shows the effectiveness of the algorithm in minimizing $\DP$ and, in many cases, nearly eliminating bias. Due to space limitations, we display a subset of results; similar findings for other combinations are available in our repository. \ignore{For \FodorZag, as shown in Figure~\ref{fig:comparedDSPdpD}, the reduction in bias is smaller due to an initially low bias, which limits further reduction. Nevertheless, the calibration algorithm significantly reduces the existing bias across methods. }Overall, this calibration approach effectively reduces $\DP$ across all tested algorithms and datasets.

% \begin{figure}[ht]
%     \centering
%     \subfloat[\amzgog]{\label{fig:comparedDSPdpA}
%     \includegraphics[width=0.23\textwidth]{fig/DP_Amazon-Google.pdf}}
%     \subfloat[\DBLPACM]{\label{fig:comparedDSPdpB}
%         \includegraphics[width=0.23\textwidth]{fig/DP_DBLP-ACM.pdf}}
%         % \vspace{3mm}
%     %     \subfloat[\Beer]{\label{fig:comparedDSPdpC}
%     %     \includegraphics[width=0.23\textwidth]{fig/DP_Beer.pdf}}
%     % \subfloat[\FodorZag]{\label{fig:comparedDSPdpD}
%     %     \includegraphics[width=0.23\textwidth]{fig/DP_Fodors-Zagat.pdf}}
%     \caption{Comparison of distributional demographic disparity across different models and datasets before and after \calib.}
%     \label{fig:comparedDSPdp}
% \end{figure}

\subsubsection{Bias Reduction in Conditional Calibration} \label{sec:exp-con-calibrate}
In this section, we first show that \calib is not suitable for addressing biases beyond $\DP$. This is because reducing $\DP$ only requires similar score distributions between subgroups, which is insufficient to satisfy fairness criteria such as $\EO$. To overcome this, we use our alternative solution, \ccalib. In this section, we again use the entire test dataset as the input dataset $D$.

\begin{figure}[h!]
    \centering
        \centering
        \subfloat[\amzgog]{\label{fig:comparedDSPdpA}
            \includegraphics[width=0.45\textwidth]{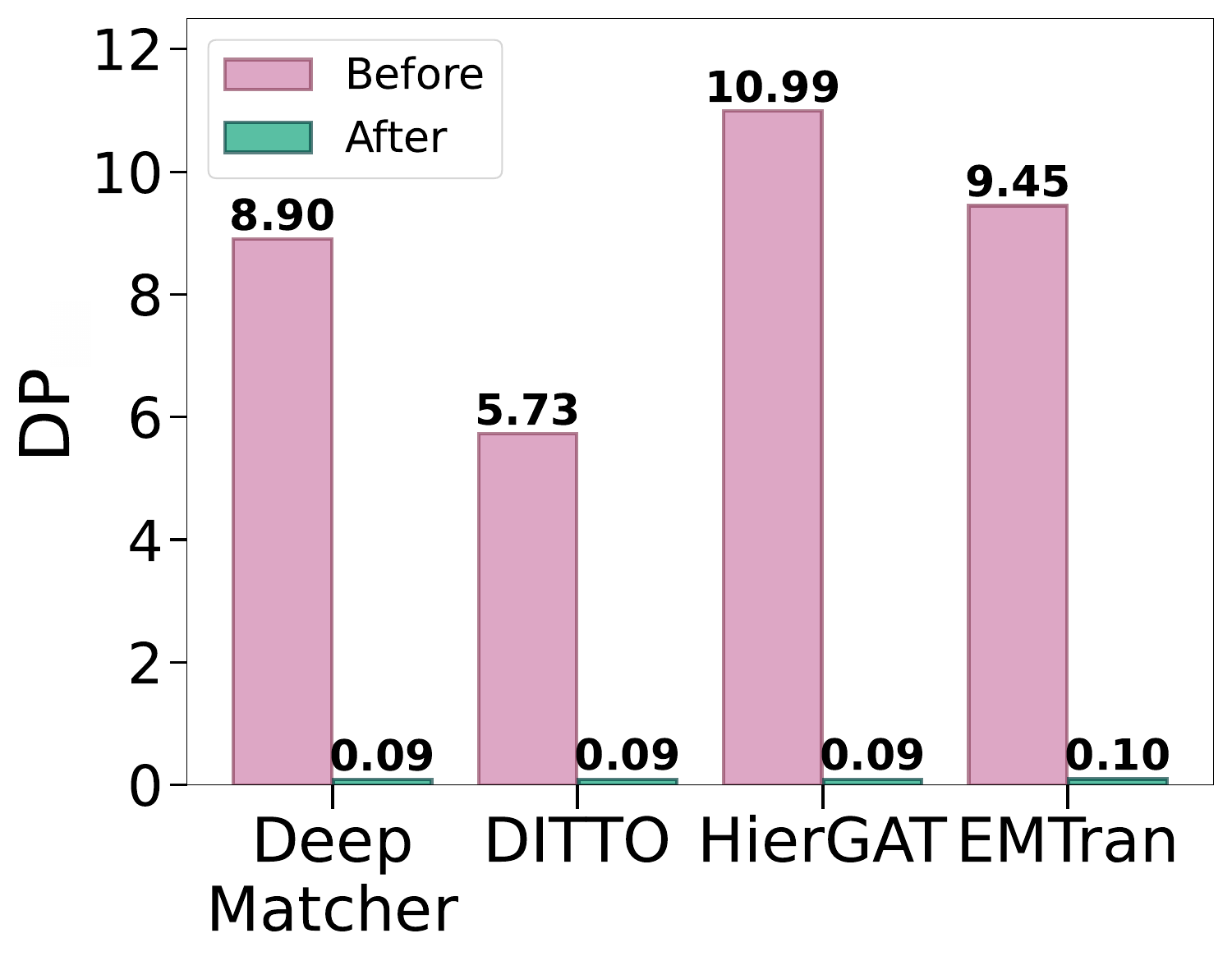}}
        \hfill
        \subfloat[\DBLPACM]{\label{fig:comparedDSPdpB}
            \includegraphics[width=0.45\textwidth]{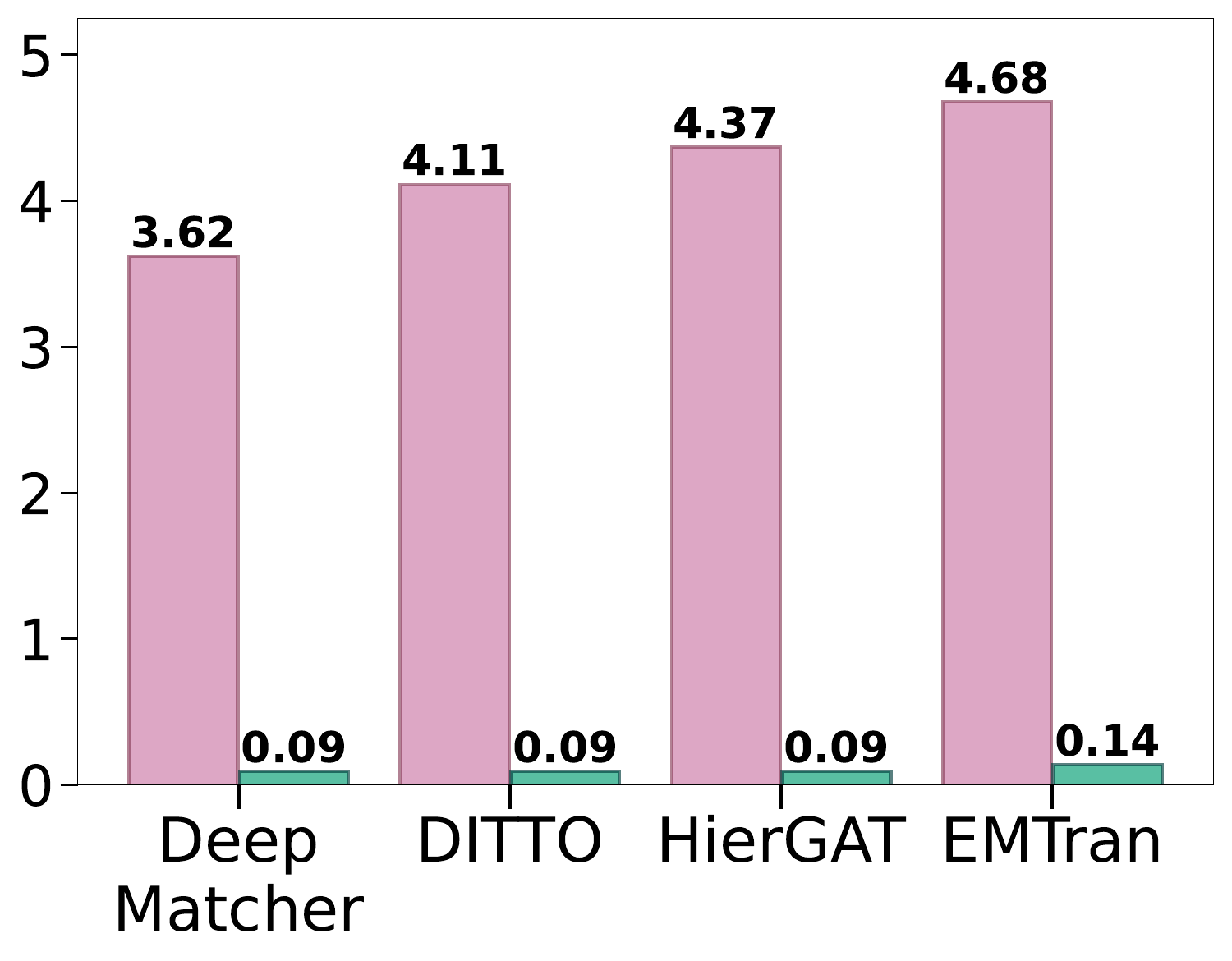}}
        \captionof{figure}{Comparison of distributional demographic disparity across different models and datasets before and after \calib.}
        \label{fig:comparedDSPdp}
    
        \centering
        \subfloat[$\amzgog$]{\label{fig:comparedEOeodA}
            \includegraphics[width=0.45\textwidth]{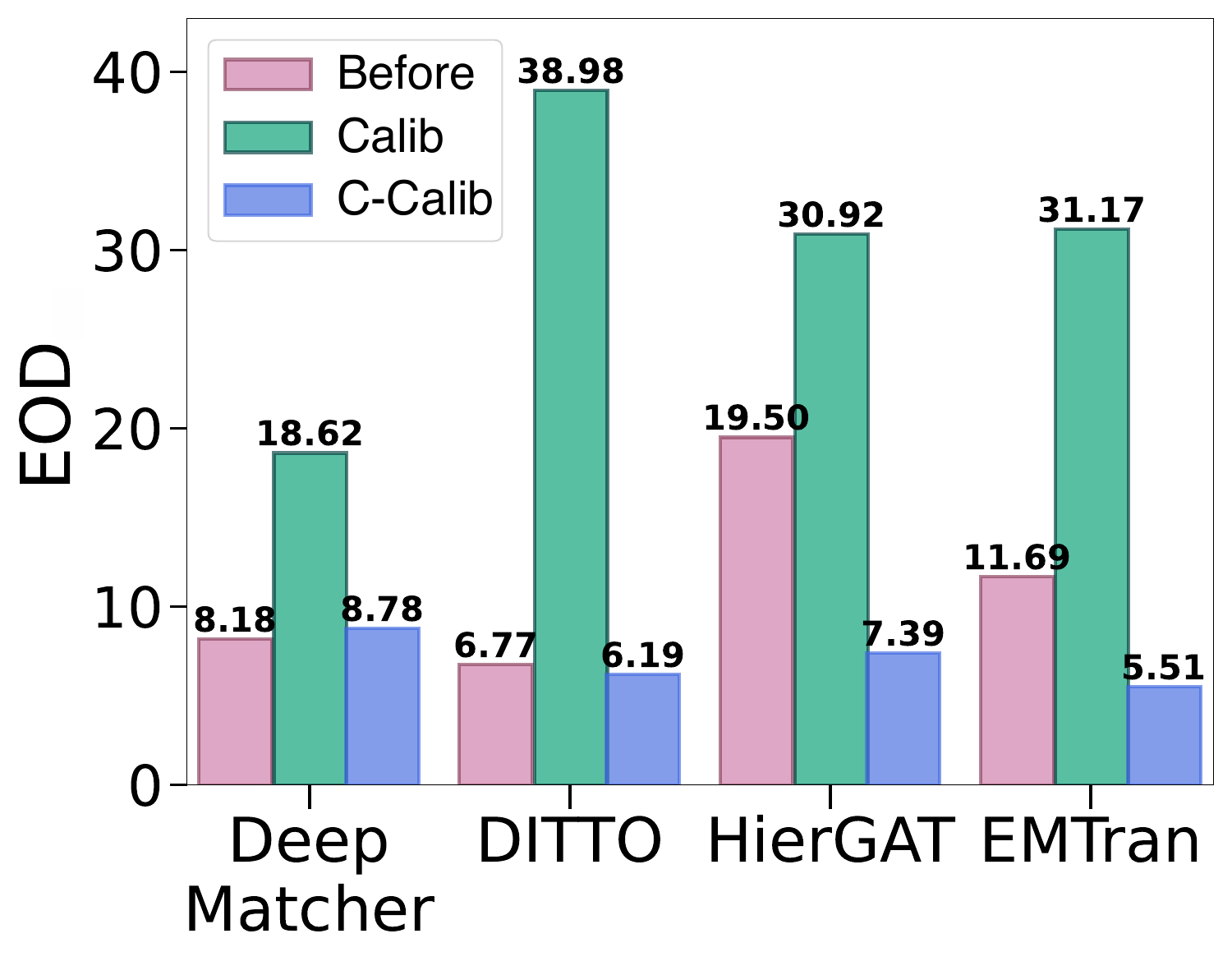}}
        \hfill
        \subfloat[$\Beer$]{\label{fig:comparedEOeodC}
            \includegraphics[width=0.45\textwidth]{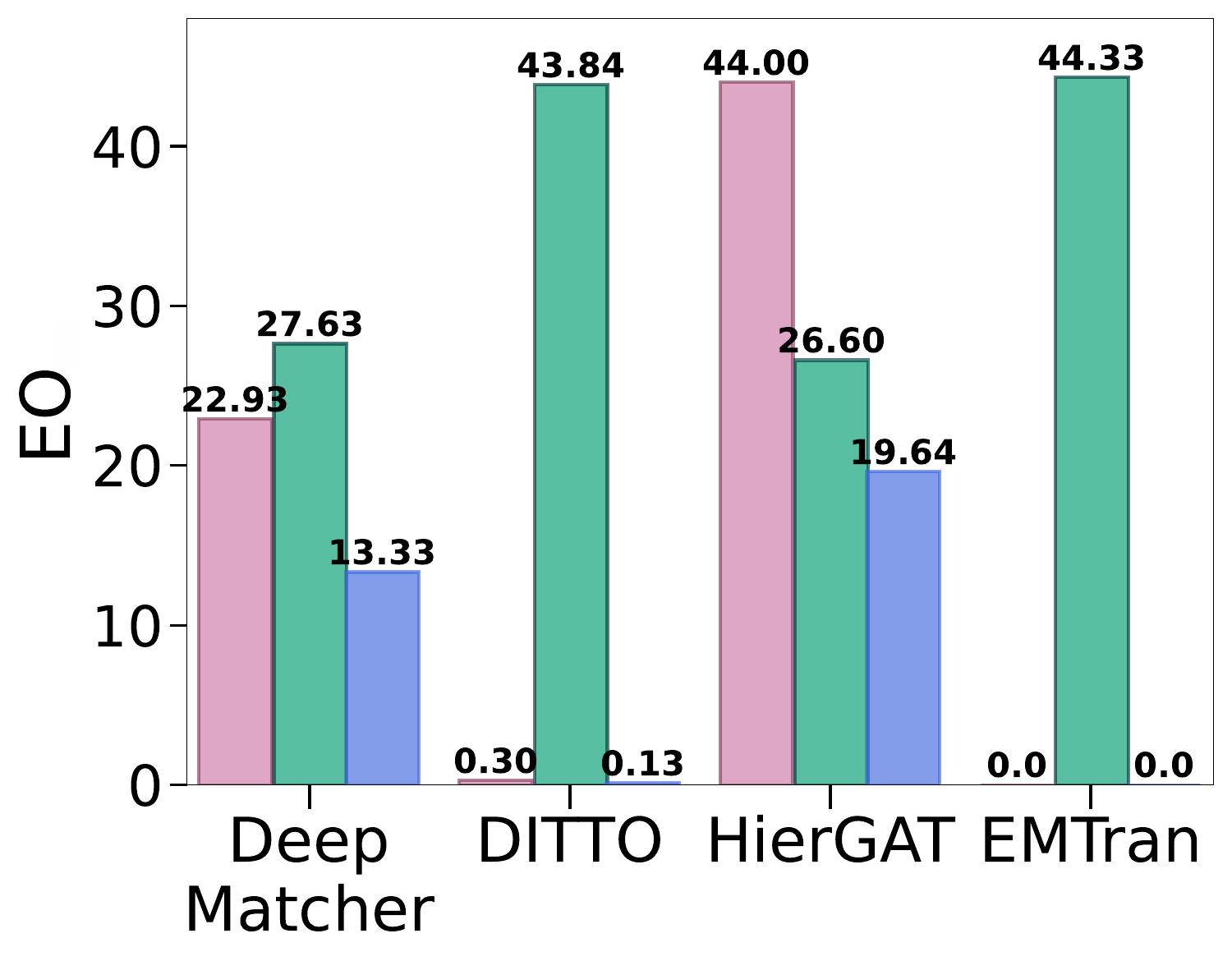}}
        \captionof{figure}{Comparison of $\EO$ and $\EOD$ across different models and datasets before and after \calib and \ccalib.}
        \label{fig:comparedEOeod}    
\end{figure}

Figure~\ref{fig:comparedEOeod} compares the effectiveness of \calib and \ccalib in reducing $\EO$ and \EOD. As shown, \calib only reduces $\EO$ and \EOD in a few cases (e.g., \hiergat on \Beer); in most instances, it fails to reduce these biases because it does not incorporate label information.This figure demonstrates that \calib's effect on $\EO$ and \EOD is inconsistent: in some cases, it reduces bias, while in others it increases bias or has no impact. This variability arises from differences in label distributions and the matching quality of methods.

Figure~\ref{fig:comparedEOeod} also shows the results for \ccalib. As evident in this Figure, there is a notable improvement in reducing both $\EO$ and \EOD across all models and datasets compared to \calib. The conditional approach is more effective in reducing these fairness metrics, which aligns with its design of adjusting score distributions conditioned on estimated labels.
For instance, \ccalib on the \amzgog dataset (Figure~\ref{fig:comparedEOeodA}) reduced \EOD more effectively across all models compared to \calib. Specifically, \hiergat's \EOD for \amzgog was increased to 30.92\% from its initial value of 19.50\% by \calib; however, \ccalib reduced it to 7.39\%. Similarly, on the \Beer dataset (Figure~\ref{fig:comparedEOeodC}), the $\EO$ for \ditto and \hiergat is significantly reduced to near-zero values after \ccalib, indicating a strong improvement over \calib. Notably, for \ditto on \Beer in Figure~\ref{fig:comparedEOeodC}, the initial $\EO$ was 0.30\%, which is already minimal, but \ccalib further reduced it to 0.13\%, highlighting the effectiveness of this method.
Across other datasets (available in our repository), this pattern of reduced disparity by \ccalib continues, showing that the conditional approach better aligns score distributions across demographic groups by estimating labels.\ignore{, thus achieving enhanced fairness in both \EO and \EOD across thresholds. }

\begin{figure}[!t]
    \centering
    \subfloat[$\Delta$\EO using \calib]{\label{fig:calibrateExamplealg1EO}
        \includegraphics[width=0.45\linewidth]{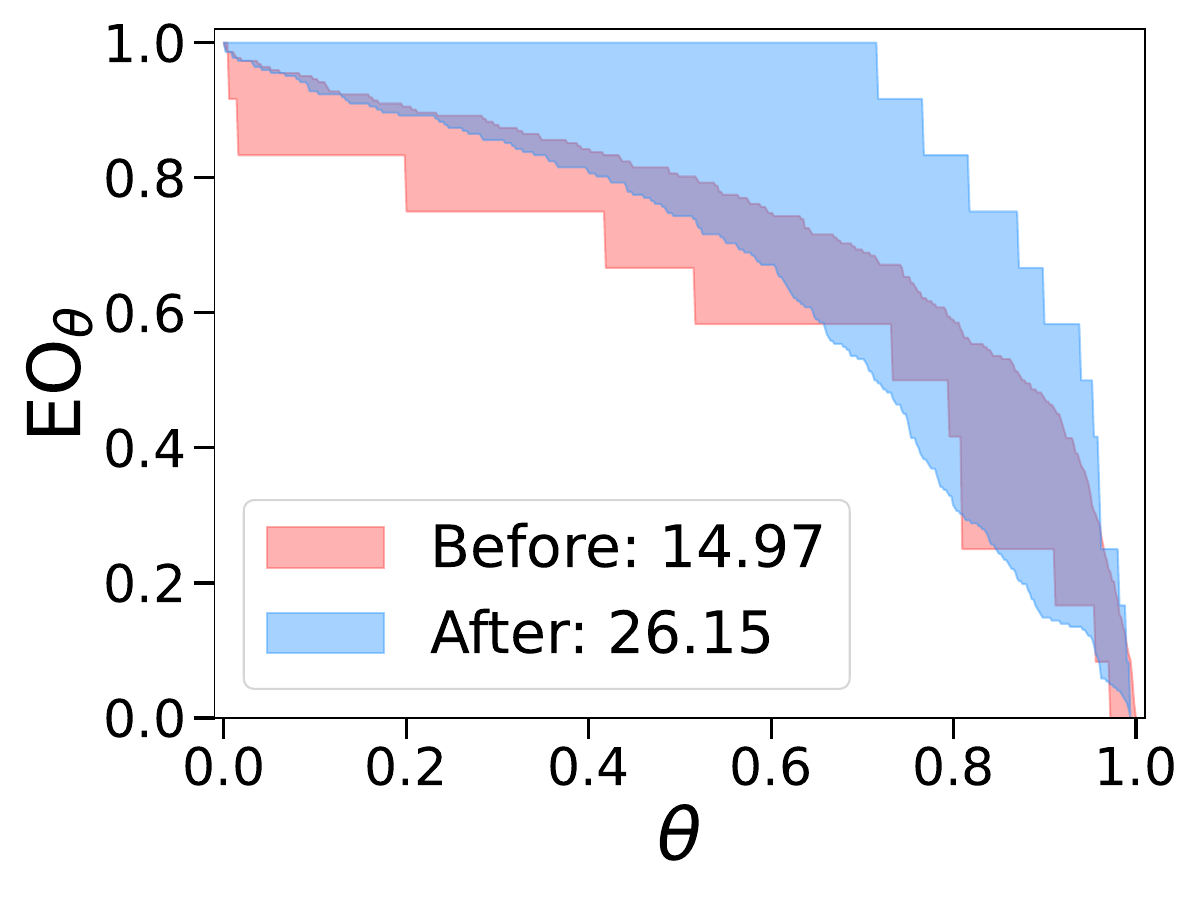}}
    \subfloat[$\Delta$\EOD using \calib]{\label{fig:calibrateExamplealg1EOD}
        \includegraphics[width=0.45\linewidth]{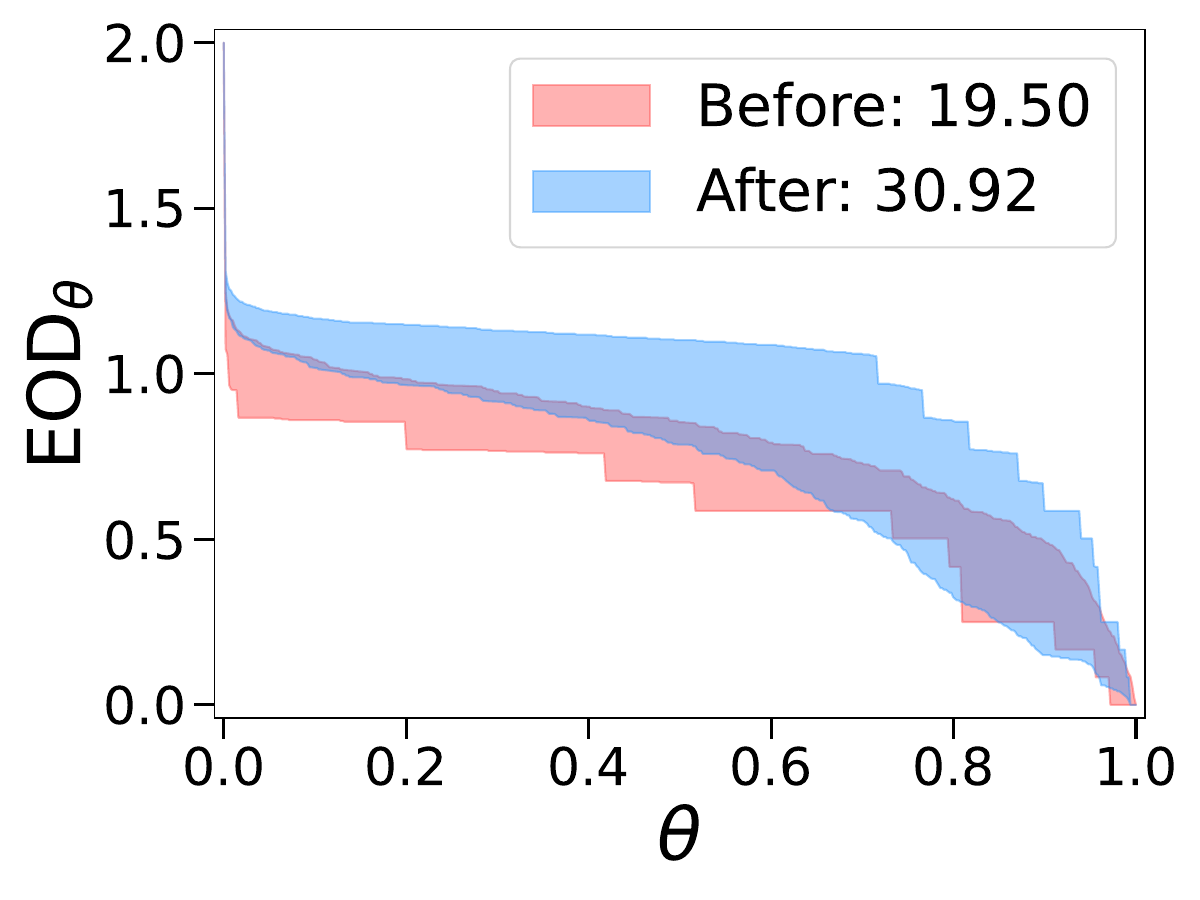}}

    \subfloat[$\Delta$\EO using \ccalib]{\label{fig:calibrateExamplealg2EO}
        \includegraphics[width=0.45\linewidth]{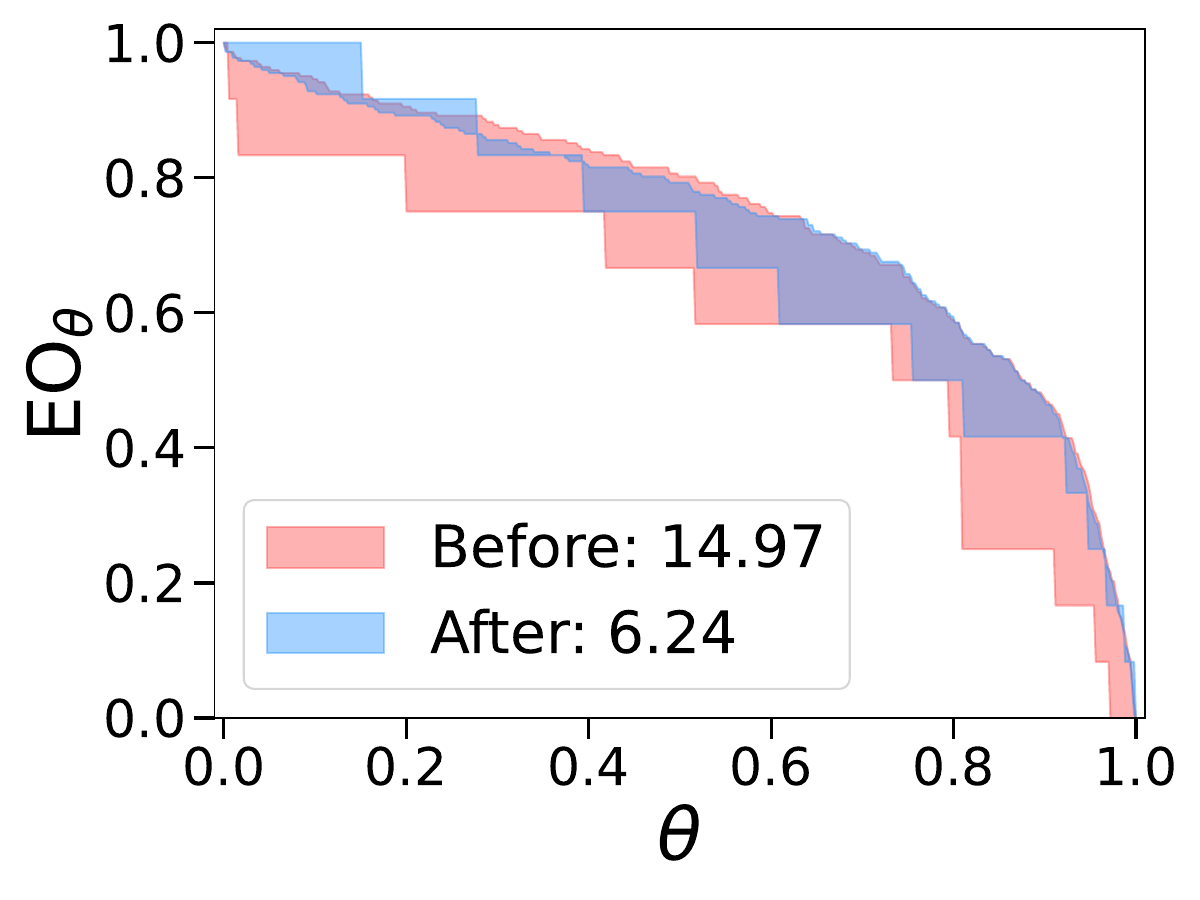}}
    \subfloat[$\Delta$\EOD using \ccalib]{\label{fig:calibrateExamplealg2EOD}
        \includegraphics[width=0.45\linewidth]{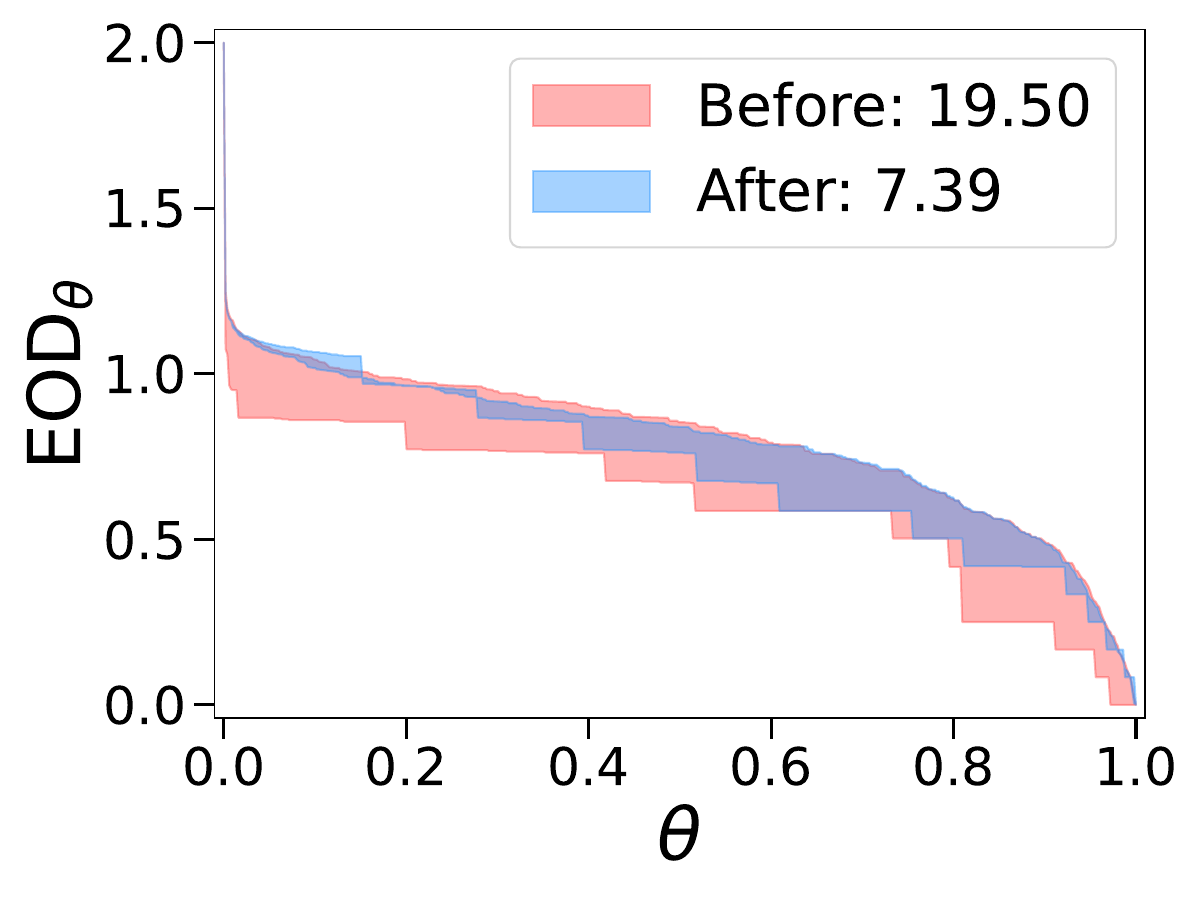}}

    \caption{Comparison of \EO and \EOD differences, before and after calibration across various thresholds for \hiergat on \amzgog dataset using \calib and \ccalib.}
    \label{fig:calibrateExamplealg1}
\end{figure}

To analyze why \calib fails to reduce certain biases, we present Figure~\ref{fig:calibrateExamplealg1}. This figure illustrates \hiergat on \amzgog dataset, showing \EO and \EOD across thresholds before and after calibration. In each plot, the width at each threshold shows the bias magnitude at that threshold.
Figures~\ref{fig:calibrateExamplealg1EO} and~\ref{fig:calibrateExamplealg1EOD} show that after applying \calib, $\EO$ increases from 14.97\% to 26.15\% and \EOD from 19.50\% to 30.92\%. This highlights \calib’s limitation in addressing these metrics, as it does not consider labels.
Additionally, before calibration, minority group exhibits lower values compared to majority group (visible as a more discrete edge in the plot due to the minority's smaller sample size). Post-calibration, values for majority group drop while those for minority group increase.
This suggests that aligning score distributions with targeted adjustments for minority group might help address $\EO$ and \EOD, however it would require changes beyond \calib.

Similarly, Figures ~\ref{fig:calibrateExamplealg2EO} and ~\ref{fig:calibrateExamplealg2EOD} shows the effect of \ccalib in the same setting. By comparing this figures with the performance of \calib on this model and dataset, it is clear that \ccalib performs substantially better at reducing \EO and \EOD biases. Here, \ccalib reduces \EO bias from 14.97\% to 6.24\% and \EOD from 19.50\% to 7.39\%. This reduction shows the effectiveness of \ccalib in aligning score distributions conditioned on estimated labels and achieving improved fairness across thresholds. It is also evident that \EO and \EOD biases are reduced almost consistently across all thresholds.

Furthermore, by examining the edge for the minority group (a more step-like edge) in Figure~\ref{fig:calibrateExamplealg2EO} and Figure~\ref{fig:calibrateExamplealg2EOD}, we observe that the degree of change in scores for both groups is controlled, causing the two edges to move closer together without crossing each other. This results in a reduction in bias. 
As a result, this algorithm effectively controls the magnitude of score adjustments, which helps mitigate bias.

The main takeaway here is that \calib is highly effective in reducing \DP but is not suitable for reducing \EO and \EOD biases. In contrast, \ccalib provides a more effective bias reduction approach for \EO and \EOD compared to \calib.

\subsubsection{Risk of Calibration}\label{sec:exp-risk}
To conclude our analysis, we examine how \calib and \ccalib affect performance preservation after calibration. Definition~\ref{df:fair-problem} defines \fRisk as the expected absolute change in scores, which directly quantifies how much calibration perturbs the original matcher. In this section, however, we do not directly estimate this score-space risk; instead, we report the change in AUC as an application-facing, threshold-independent proxy for utility. A small AUC change indicates that the calibrated scores largely preserve the ranking quality of the original matcher.

We emphasize that AUC change is not theoretically equivalent to \fRisk in general: small score perturbations can sometimes alter rankings, and conversely some score changes may leave rankings (and hence AUC) unchanged. Thus, our AUC-based analysis should be interpreted as measuring \emph{utility impact} (performance preservation), complementary to the formal risk notion in Definition~\ref{df:fair-problem}. Establishing tighter theoretical connections between score deviation and AUC preservation is an interesting direction for future work.

Table~\ref{tab:comparedDSPdpAUCalg2} shows the effect of \calib\ and \ccalib\ on AUC for \amzgog\ and \DBLPGogS\ datasets across matching models. For clarity, only the AUC decreases after calibration are reported.
In all cases, both algorithms have minimal AUC reductions, and similar results are observed universally, however only a subset of results is reported due to space constraints. 
In \amzgog dataset, AUC decrease post-calibration using \calib at most 1.01\%. Note that a maximum reduction of 1.01\% in AUC results in almost completely removing $\DP$ bias from an initially significant value. Similarly, in \DBLPGogS dataset, there is almost no change in AUC after calibration using \calib across different matchers, indicating that \calib\ effectively reduces \(\DP\) without compromising accuracy.

Another important observation in this table is that \ccalib achieves slightly better stability in AUC compared with \calib. For instance, in \amzgog, \ditto's AUC decreases only 0.28\%, a minor difference compared to the larger drop of 1.01\% for \calib. This trend is similar across other models and datasets, and sometimes, the AUC shows almost no change for \ccalib. This improvement suggests that \ccalib better preserves model performance, which can be attributed to its consideration estimated labels during calibration.

Overall, these results show that both \calib and \ccalib achieves their fairness objective, reducing bias while minimizing risk, and even when the initial bias is significant, the reduction in AUC does not exceed 1\%, which is impressive.

% \begin{figure}[ht]
%     \centering
%     \subfloat[\amzgog]{ \label{fig:comparedDSPdpAUCalg2A}
%     \includegraphics[width=0.245\textwidth]{fig/AUC_Amazon-Google_alg2.pdf}}
%     \subfloat[\DBLPGogS]{\label{fig:comparedDSPdpAUCalg2B}
%         \includegraphics[width=0.245\textwidth]{fig/AUC_DBLP-GoogleScholar_alg2.pdf}}
%     \caption{Comparison of AUC across different models and datasets before and after \calib and \ccalib}
%     \label{fig:comparedDSPdpAUCalg2}
% \end{figure}

\begin{table}[ht]
\centering
    \renewcommand{\arraystretch}{1} % Adjust row height
    \setlength{\tabcolsep}{3mm} % Adjust global column spacing
    \resizebox{\textwidth}{!}{
    \normalsize

\begin{tabular}{lccc|ccc}
\toprule
\multirow{2}{*}{\textbf{Model}} & \multicolumn{3}{c|}{\textbf{\amzgog}} & \multicolumn{3}{c}{\textbf{\DBLPGogS}} \\
\cmidrule{2-7}
 & \textbf{\texttt{Before}} & \textbf{\calib} & \textbf{\ccalib} & \textbf{\texttt{Before}} & \textbf{\calib} & \textbf{\ccalib} \\
\midrule
\ditto   & 96.72 & -1.01 & -0.28 & 99.70 & -0.11 & -0.01 \\
\hiergat & 96.93 & -0.99 & -0.11 & 99.72 & -0.10 & -0.01 \\
\EMT  & 96.15 & -0.97 & -0.24 & 98.72 & -0.05 & -0.01 \\
\hmatcher  & 90.06 & -1.01 & -0.06 & 99.34 & -0.10 & 0.0 \\
\bottomrule
\end{tabular}
}
\caption{Comparison of AUC.
{\em Before} is the baseline AUC prior to calibration. \calib and \ccalib columns show \emph{changes} in AUC after each calibration algorithm.
}
\label{tab:comparedDSPdpAUCalg2}
\end{table}

\subsubsection{Role of Input Dataset $D$}\label{sec:exp-dataseteffect}
In this section, we evaluate the effect of selecting different input datasets $D$ for the \calib algorithm. Similar trends were observed for the \ccalib algorithm. Table~\ref{tab:datasetDcompare12} shows the results of applying two matching models on four benchmark datasets, comparing two calibration settings: first using the same set of query points as the input dataset $D$, and second using a different dataset 
$D$, specifically the training data used for learning the matching models.

As shown in Table~\ref{tab:datasetDcompare12}, when the input dataset $D$ matches the set of query points, the calibration almost completely eliminates \DP. However, when using the training data as $D$, \DP is still reduced but not as effectively. This is because, although the training and test sets are drawn from similar distributions, they are not perfectly aligned. In fact, choosing an arbitrary dataset as 
$D$ can even increase \DP after calibration. Therefore, careful selection of dataset $D$ is crucial.

\begin{table}[ht]
\centering
    \renewcommand{\arraystretch}{1} % Adjust row height
    \setlength{\tabcolsep}{3mm} % Adjust global column spacing
    \resizebox{\textwidth}{!}{
    \normalsize

\begin{tabular}{lccc|ccc}
\toprule
\multirow{2}{*}{\textbf{Dataset}} & \multicolumn{3}{c|}{\textbf{\dmatcher}} & \multicolumn{3}{c}{\textbf{\ditto}} \\
\cmidrule{2-7}
 & \textbf{\texttt{Before}} & \textbf{$D$: Test} & \textbf{$D$: Train} & \textbf{\texttt{Before}} & \textbf{$D$: Test} & \textbf{$D$: Train} \\
\midrule
\amzgog   & 8.90 & 0.09 & 1.13 & 5.73 & 0.09 & 1.67 \\
\DBLPACM  & 3.62 & 0.09 & 0.58 & 4.11 & 0.09 & 1.90 \\
\DBLPGogS & 6.15 & 0.07 & 1.71 & 5.8 & 0.07 & 2.68 \\
\itunamz  & 13.84 & 0.65 & 13.80 & 10.75 & 0.65 & 13.89 \\

\bottomrule
\end{tabular}
}
\caption{
\DP (\%) before and after calibration using different input datasets \(D\). 
``$D$: Test'' refers to using the same set of query points (i.e., test data) for calibration, while ``$D$: Train'' uses the training data that was used to train the matcher.
}
\label{tab:datasetDcompare12}
\end{table}

\section{Related Work} \label{sec:rw}

This section reviews fairness literature in record matching, ranking, and regression, highlighting connections to our work.

\subsection{Fairness in Record Matching}

% Existing research on fairness in record Matching has mainly treated matching as a binary classification task~\cite{efthymiou2021fairer, shahbazi2023through, yang2023minimax}. In these studies, fairness is evaluated based on binary outcomes without considering potential biases in the underlying matching scores. 
% The closest prior research to our work proposes a threshold-independent fairness metric based on the AUC~\cite{nilforoushan2022entity}. Although threshold-independent, the \AUC-based metric may not fully capture bias and can be misleading in some cases~\cite{kwegyir2023misuse}.
% Other than defining fairness metrics that are threshold agnostic, some researchers have attempted to propose bias reduction methods for record matching. For instance,~\cite{makri2022towards} proposed a training approach where separate models are trained for different groups to improve both fairness and accuracy. In a related work,~\cite{efthymiou2021fairer} extended matching models by embedding fairness constraints directly into the matching process. Additionally,~\cite{ebraheem2018distributed} created distributed representations of tuples, focusing on accuracy while adjusting the embeddings to maintain fairness. Tools for auditing matching algorithms for bias at various stages have also been developed~\cite{2024arXiv240407354S}, incorporating fairness checks during both training and testing.
% Furthermore, \cite{shahbazi2023through} provide a comprehensive survey on fairness in matchers by reviewing various models and applying traditional fairness metrics. 

Most existing work on fairness in record matching treats it as a binary classification task~\cite{efthymiou2021fairer, shahbazi2023through, yang2023minimax}, evaluating fairness only through binary outcomes without addressing potential biases in matching scores. The most relevant prior work introduces a threshold-independent fairness metric based on AUC~\cite{nilforoushan2022entity}, but this can still be misleading in some cases~\cite{kwegyir2023misuse}. Beyond defining such metrics, some studies propose bias reduction methods. For example, \cite{makri2022towards} train separate models per group, while \cite{efthymiou2021fairer} embed fairness constraints into the matching process. Others, like~\cite{ebraheem2018distributed}, adjust tuple embeddings to preserve fairness alongside accuracy. Auditing tools have also been developed~\cite{2024arXiv240407354S} to check fairness during training and testing. Finally,~\cite{shahbazi2023through} provide a survey of fairness in matchers using standard fairness metrics.

\subsection{Fairness in Ranking}\label{sec:fairrankinglit}
Fair ranking aims to reduce bias when ordering items, typically by selecting $k$ candidates from $n$ inputs. Methods are either score-based, assigning scores to rank items, or supervised, directly selecting the top-$k$ without scoring~\cite{10.1145/3533379}. While both fair ranking and score-based matching rely on scores, their goals differ: ranking emphasizes fair item positions, while matching focuses on producing unbiased scores without missing true matches. Our work treats matching as binary classification, where scores are intermediate values--similar to a $k'$--ranking with an unknown number of match pairs. Unlike ranking, which emphasizes proportional inclusion, we aim to align score distributions across groups to ensure fairness at all decision thresholds, as measured by metrics like demographic parity.

% Fair ranking aims to reduce bias when ordering items, typically selecting $k$ candidates from $n$ inputs. Ranking methods are divided into score-based approaches, where items receive scores for ranking, and supervised approaches, which directly select the top-$k$ items without scoring~\cite{10.1145/3533379}.
% While fair ranking and score-based matching both rely on scores, their goals differ: ranking emphasizes fair item positions, whereas fairness in matching ensures unbiased scores without losing true matches. While ranking aims to fairly select the top-$k$ candidates, we focus on binary classification, treating scores as intermediate values. This is similar to a $k'$-ranking, where $k'$ is the unknown number of ``match'' pairs. Our goal is to align score distributions across groups for fairness at all decision thresholds, rather than to produce fair rankings. In ranking, the goal is balanced group representation through proportional inclusion, however fairness metrics like demographic parity focus on equalizing score distributions regardless of rank.

The work by~\cite{10.1145/3533379} surveys fairness in score-based ranking, explaining how biased scores can cause unfair rankings. Similarly,~\cite{10.1145/3533380} reviews fairness in supervised ranking, showing how bias in training data leads to unfair outcomes. Since supervised ranking does not use scores, post-processing adjustments are generally not applicable for it.
In another work,~\cite{zehlike2017fa} ensures protected groups are fairly represented in the top-$k$ results through constrained optimization. Additionally,~\cite{chakraborty2022fair} focuses on fair rank aggregation by applying proportional representation while minimizing changes to the original rankings. Also,~\cite{singh2019policy} introduces a policy method that balances individual and group fairness when learning ranking policies. Moreover, ~\cite{pitoura2022fairness} provides a broad overview of fairness in ranking and recommendation systems, focusing on equitable visibility. Furthermore, ~\cite{yang2019balanced} incorporates diversity constraints into ranking optimization to achieve balanced group representation. 
Finally,~\cite{celis2018ranking} formulates ranking as an optimization problem with fairness constraints, balancing fairness and relevance.

\subsection{Fairness in Regression} \label{sec:fairreglit}

Fairness-aware regression aims to predict continuous outcomes while reducing bias. This is closely related to fairness in record matching scores, as both involve assigning continuous values. 
In both cases, fairness focuses on aligning the distributions of these continuous outputs across demographic groups to prevent systematic differences. However, they have differences.
Matching scores is used to decide if two records refer to the same entity, so fairness needs to ensure that scores are unbiased without harming the ability to detect true matches. In contrast, fair regression focuses on making accurate and unbiased predictions for individual outcomes. This makes fairness in matching more sensitive, as changing scores can hurt matching accuracy, making the trade-off between fairness and performance more challenging.

Many methods have been proposed to address bias in regression. For example,~\cite{agarwal2019fair} transformed regression into a weighted classification problem. This approach adds fairness constraints into the optimization, balancing fairness with accuracy. Similarly,~\cite{chzhen2020fair} introduces a method that uses Wasserstein barycenters to align the distribution of predictions across demographic groups, reducing bias while staying close to the original data.
Additionally,~\cite{komiyama2018general} proposed a framework for fair regression by formulating fairness as a constrained optimization problem. This allows for different fairness goals, such as demographic parity and individual fairness, to be applied during training. In contrast,~\cite{kwegyir2023repairing} focuses on fixing fairness issues after training by adjusting the regression outputs to ensure fair binary decisions at any threshold. \ignore{This is similar to post-processing calibration in matching, where scores are adjusted to meet fairness criteria without retraining the original model.}
Finally, in another study by~\cite{fukuchi2024demographic}, a minimax optimization framework is presented to enforce demographic parity in regression while preserving accuracy.

\section{Conclusion and Future Work} \label{sec:conclusionfinal}

This work focuses on biases in matching scores used in record matching, which cannot be detected or corrected using traditional fairness metrics designed for binary classification. Unlike binary labels, matching scores are used across various thresholds and directly influence downstream decisions such as ranking and human review. We show that existing state-of-the-art models can produce biased score distributions, even when they appear fair at specific thresholds. To address this, we introduce a threshold-independent measure of bias based on the statistical divergence between score distributions across demographic groups. Our evaluation demonstrates that this metric reveals biases overlooked by existing approaches and better captures fairness concerns in score-based systems.

To mitigate these biases, we propose two post-processing calibration algorithms that modify matching scores without altering the underlying models or requiring access to their training data. The first method, \calib, aligns score distributions across groups using the Wasserstein barycenter, effectively reducing \DP bias. The second method, \ccalib, extends this idea to account for true labels, achieving fairness criteria such as \EO and \EOD. Both algorithms are model-agnostic and offer theoretical guarantees on fairness and risk. Empirical results across a range of datasets and models show that our methods significantly reduce bias while preserving predictive performance, offering a practical and principled approach for fairer record matching.

Future research could extend this work in several directions. Expanding calibration techniques to address additional fairness metrics beyond \DP, \EO, and \EOD would make this approach more versatile. Exploring pre-processing and in-processing methods alongside post-processing calibration could provide a more complete understanding of the trade-offs between fairness, accuracy, and computational efficiency. Another valuable avenue involves examining biases throughout the entire record matching pipeline, including the blocking, matching, and resolution stages. An end-to-end analysis of how biases propagate and interact across these stages could lead to more comprehensive and fairness-aware entity resolution methods. Our calibration-based approach provides a strong foundation for improving fairness in record matching. Further exploration of extended bias metrics, pipeline-level bias analysis, and comparisons of different mitigation techniques will help build more fair and robust systems for record matching.

\bibliographystyle{elsarticle-num}
\bibliography{ref}

\vspace{-1.8cm}

% \begin{IEEEbiography}[{\includegraphics[width=1in,height=1.25in,clip,keepaspectratio]{fig/tmp.jpg}}]{Mohammad Hossein Moslemi} Mohammad Hossein Moslemi
% is a Master's student in Computer Science at Western University, Canada. His research focuses on responsible AI, algorithmic fairness, and optimal transport applications in data management. He has presented his work at SIGMOD and IEEE Big Data and has contributed to fairness-aware entity matching and data cleaning. His work aims to improve quality, fairness, and interpretability in data-driven systems.
% \end{IEEEbiography}

% \vspace{-2cm}

% \begin{IEEEbiography}[{\includegraphics[width=1in,height=1.25in,clip,keepaspectratio]{fig/tmp.jpg}}]{Mostafa Milani}
% Mostafa Milani is an Assistant Professor in the Department of Computer Science at Western University, Canada. His research focuses on data management and databases, data quality and cleaning, data privacy, and responsible and fair data management. He has published in top-tier venues such as SIGMOD, VLDB, ICDE, and TKDE and serves regularly on their program committees. His work aims to improve quality, fairness, privacy, and interpretability in data-driven systems.
% \end{IEEEbiography}

% \vfill

%\input{appendix}

\end{document}